\documentclass[review]{elsarticle}

\def\mycopyrightnotice{
  {\footnotesize
  \begin{minipage}{\textwidth}
    First publication in the ``Cahiers du GERAD'' series in February 2020. \\
    Submitted to EURO Journal on Transportation and Logistics on January 17, 2020. \\
    Accepted August 24, 2020, available online September 2, 2020. \\
    \url{https://doi.org/10.1016/j.ejtl.2020.100020}
  \end{minipage}
  }}

\usepackage{lipsum}
\makeatletter
\def\ps@pprintTitle{%
\let\@oddhead\@empty
\let\@evenhead\@empty
\def\@oddfoot{\mycopyrightnotice}
\let\@evenfoot\@oddfoot}
\makeatother

\journal{EURO Journal on Transportation and Logistics}

\usepackage{natbib}

\usepackage{tikz}
\usetikzlibrary{calc, positioning, shapes, fit, tikzmark}
\usepackage{color,soulutf8,longtable,colortbl,setspace,ifthen,xspace,url,pdflscape}
\usepackage[justification=centering]{caption}
\usepackage{amsmath}

\usepackage[normalem]{ulem}
\usepackage{booktabs}

\usepackage{array}
\newcolumntype{L}[1]{>{\raggedright\let\newline\\\arraybackslash\hspace{0pt}}m{#1}}
\newcolumntype{C}[1]{>{\centering\let\newline\\\arraybackslash\hspace{0pt}}m{#1}}
\newcolumntype{R}[1]{>{\raggedleft\let\newline\\\arraybackslash\hspace{0pt}}m{#1}}

\usepackage{float}
\usepackage{amsfonts}

\usepackage{stmaryrd}

\usepackage{hyperref}       
\hypersetup{ %
	pdftitle={Machine Learning in Airline Crew Pairing to Construct Initial Clusters for Dynamic Constraint Aggregation},
	pdfkeywords={},
	pdfborder=0 0 0,
	pdfpagemode=UseNone,
	colorlinks=true,
	linkcolor=blue, 
	citecolor=blue, 
	filecolor=blue, 
	urlcolor=blue, 
	pdfview=FitH,
	pdfauthor={Anonymous},
}

\begin{document}

\begin{frontmatter}

\title{Machine Learning in Airline Crew Pairing to Construct Initial Clusters for Dynamic Constraint Aggregation}

\author[address1,address2,address4]{Yassine Yaakoubi\corref{mycorrespondingauthor}}
\cortext[mycorrespondingauthor]{Corresponding author}
\ead{yassine.yaakoubi@polymtl.ca}
\author[address1,address2]{Fran\c{c}ois Soumis}
\ead{francois.soumis@gerad.ca}
\author[address3,address4]{Simon Lacoste-Julien}
\ead{slacoste@iro.umontreal.ca}

\address[address1]{Department of Mathematics and Industrial Engineering, Polytechnique Montr\'eal, Canada}
\address[address2]{GERAD, Montr\'eal, Canada}
\address[address3]{Department of Computer Science and Operations Research, University of Montr\'eal, Canada}
\address[address4]{MILA, Montr\'eal, Canada}

\begin{abstract}
The crew pairing problem (CPP) is generally modelled as a set partitioning problem where the flights have to be partitioned in pairings. A pairing is a sequence of flight legs separated by connection time and rest periods that starts and ends at the same base. Because of the extensive list of complex rules and regulations, determining whether a sequence of flights constitutes a feasible pairing can be quite difficult by itself, making CPP one of the hardest of the airline planning problems. In this paper, we first propose to improve the prototype \textit{Baseline} solver of Desaulniers et al. \cite{desaulniers2019} by adding dynamic control strategies to obtain an efficient solver for large-scale CPPs: Commercial-GENCOL-DCA. These solvers are designed to aggregate the flights covering constraints to reduce the size of the problem. Then, we use machine learning (ML) to produce clusters of flights having a high probability of being performed consecutively by the same crew. The solver combines several advanced Operations Research techniques to assemble and modify these clusters, when necessary, to produce a good solution. We show, on monthly CPPs with up to 50 000 flights, that Commercial-GENCOL-DCA with clusters produced by ML-based heuristics outperforms \textit{Baseline} fed by initial clusters that are pairings of a solution obtained by rolling horizon with GENCOL. The reduction of solution cost averages between 6.8\% and 8.52\%, which is mainly due to the reduction in the cost of global constraints between 69.79\% and 78.11\%.

\end{abstract}

\begin{keyword}
Machine Learning\sep Column Generation\sep Constraint Aggregation\sep Airline Crew Scheduling\sep Crew Pairing
\end{keyword}

\end{frontmatter}

\section{Introduction}

Crew scheduling is of crucial importance to airlines, as crew costs represent their second-largest source of expenditure, after fuel costs. As the global airline industry grows in size and volume, the complexity of scheduling problems increases significantly over time. Increasing computing capacity is not enough, and innovation in algorithms is needed to solve the growing problems.

In practice, the planning processes begin with the construction of the flight schedule, a list of flights to be operated in a period of time. Then, aircraft are allocated to the scheduled flight legs (non-stop flights) based on their type and capacity in order to maximize the net profit, which is called the fleet assignment problem. The next step is to solve the aircraft routing problem where maintenance requirements of the aircraft are considered. Next, the airline must assign a cockpit and cabin crew for each of the scheduled flights, where the crew scheduling problems arise. The objective of the crew scheduling problem is to minimize crew-related costs.

Due to its complexity, the crew scheduling problem is tackled in two sequential phases, both in practice and in the literature. First, the crew pairing problem (CPP) forms a minimum-cost set of anonymous feasible pairings from the scheduled flights, such that all flights are covered exactly once, and all the pairing regulations and contractual rules are respected. Then, crew assignment combines the anonymous pairings with rest periods, vacations, preassigned activities such as training, and other breaks over a standardized month to produce a set of individual schedules for crew members.

In the context of CPP, a pairing is a sequence of flight legs that begins and ends at the same crew base (an airport where crews are stationed). A pairing contains multiple duties: sequences of flights and deadheads (repositioning) that form a day of work. Two consecutive duties inside a pairing are separated by a layover. Pairings must comply with airline regulations as well as collective agreements, such as the maximum flying time, the maximum duration of a duty, the minimum rest time in a layover, the maximum number of calendar days in a pairing, etc. The cost of a pairing approximates the crew's salary as well as other expenditures, such as hotel costs. When pilots and copilots are trained to operate on a single type of aircraft as in the considered applications, the CPP can be decomposed by aircraft type.

Since the 1990s, the most prevalent method used to solve this problem has been the branch-\&-price: column generation embedded in a branch-\&-bound structure \cite{Desrochers1989,Desaulniers1997}. Column generation is an iterative method alternating between a restricted master problem (RMP) for pairing selection and several subproblems (SP) for pairing generation. At each iteration, the RMP is simply the master problem restricted to a subset of its variables. The subproblem is a shortest path problem with resource constraints that aims at finding a feasible pairing starting on the corresponding day at the corresponding base with the least reduced cost. For more details on the method, see the survey on CPPs by Cohn and Barnhart \cite{Cohn2003}, and Deveci and Demirel \cite{Deveci2018} for a more recent survey.

In the process of developing a solver capable of dealing with much larger problems than those that have been addressed to date, GENCOL\footnote{ \url{http://www.ad-opt.com/optimization/why-optimization/column-generation/}}, a commercial optimization solver, was integrated into a rolling horizon approach. Because the original solver can only handle a few thousand flights per window, windows are constrained to be two-days long in the case of a large CPP. The length of the windows is too short to produce good solutions to monthly problems, as the constraints cover longer periods of time.

Elhallaoui et al. \cite{Elhallaoui2005} introduce the dynamic constraint aggregation (DCA) approach to reduce the number of constraints simultaneously considered in a set partitioning restricted master problem. Since DCA groups in clusters the constraints that are identical on the non-degenerate variables and keeps only one copy of the constraints in each cluster, this reduces degeneracy but also allows to cope with larger instances. The DCA solver starts with an aggregation, in clusters, of flights having a high probability of being operated consecutively by the same crew, in an optimal solution. The initial aggregation partition is produced with a heuristic solver producing pairings. It corresponds to fixing some flight-connection variables to one temporarily, which permits to replace all the flight-covering constraints of the flights in a cluster by a single constraint. DCA uses reduced costs to identify flight-connection variables that can be unfixed to improve the solution by breaking clusters. Reduced cost calculation requires dual variables for each flight covering constraint, but the aggregated problem produces only dual variables for the cluster covering constraints. Duals variables for missing constraints are obtained with a procedure solving shortest path problems. It is fast, but it does not provide the best dual solution. Elhallaoui et al. \cite{Elhallaoui_ips_2011} propose the Improved Primal Simplex (IPS), a generalization of DCA for linear programming with better mathematical bases. IPS keeps a minimum subset of independent constraints on the non-degenerate variables and uses a complementary linear programming problem (CP) to obtain the missing dual variables. This dual solution is an interior one (more central w.r.t. the dual optimal extreme points) that speeds up column generation. However, because the CP is defined with $\bar A = B^{-1}A$, the computation of $\bar A$ becomes expensive for large-scale problems.

Desaulniers et al. \citep{desaulniers2019} combine multiple methods developed and tested on small datasets in order to obtain an efficient algorithm for large-scale CPPs, namely, they combined column generation (GENCOL), DCA, multi-phase DCA (MPDCA), and IPS. This algorithm is adapted for the set partitioning problem with few supplementary linear constraints, but aggregate only set partitioning constraints permitting to use the network methods of DCA to identify the compatible and incompatible variables, where a variable is said to be compatible with respect to the partition if the flights covered by the pairing correspond to a concatenation of some clusters. Otherwise, this variable is declared incompatible. This permits the construction of the CP of IPS in less time than computing $B^{-1}A$. The CP is solved to obtain a negative reduced cost combination of incompatible variables and to obtain dual variables. These central dual solutions permit to speed up column generation compared to DCA. Furthermore, the algorithm combines some techniques of partial pricing and Branch-\&-Bound to solve large-scale problems, where DCA, MPDCA, and IPS have not been applied before: monthly CPP with complex industrial constraints. The partial pricing method MPDCA (Elhallaoui et al. \cite{Elhallaoui2008_multi}) will be explained in Section \ref{sec:Sol_Methods}, where we present the main ideas of \textit{Baseline}. We consider this algorithm as \textit{Baseline} in Section \ref{sec:imp_rolling_horizon} where we discuss the improvements, and in Section \ref{sec:Comp_Exp} where we compare the performance of our improved algorithm with respect to this \textit{Baseline}.

Since the difficulty with DCA, MPDCA, and IPS is to produce good initial clusters, \textit{Baseline} \citep{desaulniers2019} is fed with clusters from pairings of a solution obtained by rolling horizon with GENCOL. The drawback of \textit{Baseline} is that it takes more time to produce the initial clusters than to re-optimize with \textit{Baseline}. Following the flaw in this method, and with the emergence of machine learning (ML), we propose the following idea: to use ML models to find good clusters, which will be provided as initial information to OR (Operational Research) algorithms, thus improving the quality of solutions and the speed with which these solutions are discovered. Yaakoubi et al. \cite{Yaakoubi2019,yaakoubi2019these} build on this idea by studying the performance of several ML algorithms to solve the problem of the flight-connection problem, in which the objective is to predict the next flight that a crew has to follow in its schedule.

More generally, prior to Yaakoubi et al. \cite{Yaakoubi2019,yaakoubi2019these}, several other studies combined ML and OR to solve a Combinatorial Optimization (CO) problem. Indeed, in cases where there is a structural understanding of the CO problem and thus a theoretical knowledge of the decisions to be made by the optimization algorithm, ML can be used to provide quick approximations of these decisions, thus reducing the computation time required. In the literature, this is called \textit{learning by imitation}: samples of the expected behavior are available and are used in the ML model as demonstrations for learning.

There are various architectural approaches for implementing ML models for CO problems. In the simplest case, the ML model is the solver itself. This approach has been successfully used to solve some Euclidean TSP classes using deep learning. Vinyals et al. \cite{Vinyals2015} use this approach in combination with \textit{imitation learning}: using exact TSP solutions for smaller graphs and approximations for larger graphs, they generate a set of demonstrations which are then encoded by an RNN (\textit{Recurrent Neural Networks}). Another RNN serves as a decoder that can be used to produce a permutation on TSP nodes. This method creates a model capable of handling inputs of different sizes.

In more complex cases, ML is used to enrich the decision process with an existing CO algorithm. A ML model can be used to process the problem definition in order to extract a meaningful structure or to reduce the problem space. In these cases, ML will play a preparatory role for the CO algorithm. An example can be found in Kruber et al. \cite{Kruber2017} for solving mixed-integer linear program (MILP) problems. Using the Dantzig-Wolfe decomposition of a MILP instance, they train ML models to predict which of the two solvers will solve the instance optimally and faster. However, ML models are not used to find the final solutions. Similarly, Lodi et al. \cite{Lodi2019} examine the problem of locating facilities where a ML model is used to predict whether a derived problem instance will have a solution similar to its reference. They then add this additional data into the final solver, allowing for a faster computation time, thus allowing energy companies to assess the viability of facility locations without having to perform the possible but costly computation. We refer the interested reader to Bengio et al. \cite{Bengio2018} for a more detailed overview of the use of ML models in OR.

The application of ML techniques to solve CO problems has led to progress in solving airline-specific problems. The ML models used are frequently encountered in other problem areas, e.g., certain varieties of Genetic Algorithms (GA). Graf et al. \cite{Graf2016} examine flight scheduling and use traditional linear programming to solve it. The same problem is addressed in Tsai et al. \cite{Tsai2015} using a GA with chromosomes defined as two-dimensional objects representing possible schedules. Pandit et al. \cite{Pandit2018} study a new approach to aircraft fleet planning using a simple 3-layer neural network (NN). Various studies have addressed CPPs using GAs. Zeren et al. \cite{Zeren2012} introduce a "perturbation" operator for their GA. Another study on CPP using evolutionary algorithms can be found in Azadeh et al. \cite{Azadeh2013}. They implement a particle swarm optimization algorithm. Deveci and Demirel \cite{Deveci2018} address a variant of CPP, which consists of finding a set of minimum cost pairings covering all flights using various evolutionary algorithms. They have improved on previous solutions by using a hybrid algorithm (GA and hill-climbing method). Thus, their use of ML created a richer dataset for meta-heuristics. Although the proposed approach outperforms the two variants of GA presented in the paper, the instances used are small, containing up to 750 flights. Note that in all cases, ML alone cannot solve CPP: Solving a 50 000 flight-problem needs to find 50 000 crew connections. Even if ML yields a 99.9\% accuracy for each connection, the probability of finding a good feasible solution is $(.999)^{50 000} \approx 10^{-22}$.

As a first contribution, we propose Commercial-GENCOL-DCA, a new implementation of \textit{Baseline} \cite{desaulniers2019}, including new control strategies for the column generation, the constraint aggregation, and the branch-\&-bound. In particular, a dynamic control strategy is used to identify a "neighborhood" that is large enough to reach a good LP solution, but small enough to maintain a small number of fractional variables permitting to have an efficient heuristic branch-\&-bound. Note that a limited number of variables connecting consecutive flights in the current solution are unfrozen to define the neighborhood around the current solution. In addition, a variable aggregation and disaggregation strategy is proposed, reducing the size of the problem without negative aspects on the number of iterations of column generation and the complexity of branching decisions.

The second contribution is to modify the column generation solver to take advantage of an initial solution and clusters of flights with a large probability to be consecutive in a solution to speed up the column generation algorithm, where this initial information can be obtained by ML. In addition, we modify the complementary problem of IPS to reduce its density using the good information in the clusters.

As a third contribution, we show that modern ML techniques can help to automatically learn highly efficient crew pairing schedules and develop initial clusters for the solver. We use a modified version of the NNs proposed by Yaakoubi et al. \cite{Yaakoubi2019}, where, first, we use multi-layer convolutional neural networks (CNNs). Second, we use dropout (randomly drop units from the neural network during training) to prevent overfitting and batch normalization, to normalize the activations over the current batch in each hidden layer. Third, since we use a more complex ML model, we use extensive Bayesian Optimization, a  sequential  procedure  where  a probabilistic form of the model's performance is maintained using the Gaussian process.

As a fourth contribution, we detail a proof-of-concept study on a huge monthly problem with up to 50 000 flights. Starting with ML and finishing with mathematical programming will permit to solve globally larger problems and will avoid the loss of optimality due to heuristic decomposition in small time slices in the rolling horizon approach. In addition, the use of a rolling-horizon approach was improved to use clusters tailored to the flights of the current window and connecting well with the schedule of the previous window.

Note that our ML predictor does not have access to collective agreements and exact cost function from the airline company. Furthermore, it does not use airline-dependant information in any step of the learning process, as it would be a considerable amount of work to code them. In addition, it would reduce the generality of the predictor and complicate its application across different airlines.

The remainder of this article is structured as follows. A literature review on crew pairing is presented in Section \ref{sec:CP}. Section \ref{sec:Sol_Methods} presents \textit{Baseline} \cite{desaulniers2019} and the new implementation Commercial-GENCOL-DCA.
Next, Section \ref{sec:ML_Model} describes the ML predictor and introduces the cluster construction methodology, including different heuristics, to avoid constructing infeasible pairings. Section \ref{sec:Comp_Exp} reports computational results. Finally, a conclusion is drawn in Section~\ref{sec:Conclusion}.

\section{Literature review on the crew pairing problem} \label{sec:CP}

This section reviews a collection of relevant literature on the CPP to provide an overview of how researchers suggest approaching this problem. We focus on approaches developed to solve large-scale industrial problems.

According to Desaulniers et al. \cite{Desaulniers1997}, for each category of the crew and each type of aircraft fleet, CPP aims to find a set of pairings at minimal cost so that each planned flight is performed by a crew. The methodology for solving the problem depends on the size of the airline's network structure, rules, collective agreements, and cost structure~\cite{Birge2006}.

In the CPP, two flight legs can be operated by the same crew if the arrival station of the first flight leg is the same as the departure station of the second one, and the time between the flights is adequate to satisfy the crew feasibility rules. A sequence of flights forms a duty, where every two consecutive flights are separated by idle time.
The flights in a pairing are operated by a single crew, and a consecutive sequence of duty periods is referred to as a pairing, as long as the first duty period starts and the last duty period ends at the same station (base). Idle time between duty periods is called layover, and a pairing is feasible if it satisfies all safety and collective agreement rules such as:
\vspace{-0.25mm}
\begin{itemize}
\itemsep-0.25mm
\item minimum connection time between two consecutive flights;
\item maximum number of landings, maximum flying time, and maximum number of flights per duty;
\item maximum number of days and maximum number of duties in a pairing;
\item minimum rest-time between two duties and maximum span of a duty.
\end{itemize}

Finally, a set of feasible pairings constitutes a feasible solution if each scheduled flight is covered by at least one pairing. When a flight appears in more than one pairing, one crew operates the flight while the other crews are transferred between two stations for repositioning purposes. This is called deadheading. Deadheads are also used to relocate crew members either at the end of a pairing (to bring crews to base) or at the beginning of a pairing (to cover a flight departing from a non-base station).

In the literature, the CPP has been traditionally modelled as a set covering problem (SCP) or a set partitioning problem (SPP), with a covering constraint for each flight and a variable for each feasible pairing \cite{Desaulniers1997, Qiu2012, Kasirzadeh2017}.

Formally, we consider $F$ to be a set of legs that must be operated during a given period and $\Omega$ to be the set of all feasible pairings that can be used to cover these legs.
For each pairing $p \in \Omega$, let $c_p$ be its cost and $a_{fp}$, $f \in F$, be a constant equal to $1$ if it contains leg $f$ and $0$ otherwise.
Moreover, let $x_p$ be a binary variable that takes value $1$ if pairing $p$ is selected, and $0$ otherwise.
Using a set-partitioning formulation, the CPP can be modelled as follows:
\begin{align}
& \underset{x}{\text{minimize}}
& \sum_{p \in \Omega}{c_{p} x_p} \label{Eq:CPP1} \\
& \text{subject to}
& \sum_{p \in \Omega}{a_{fp} x_p} = 1
& & \forall f \in F   \label{Eq:CPP2} \\
&
& x_p \in \{ 0 , 1 \}
& & \forall p \in \Omega \label{Eq:CPP3}
\end{align}
The objective function (\ref{Eq:CPP1}) minimizes the total pairing costs. Constraints (\ref{Eq:CPP2}) ensure that each leg is
covered exactly once, and constraints (\ref{Eq:CPP3}) enforce binary requirements on the pairing variables.

Marsten et al. \cite{Marsten1981} present a first commercial system with a specialized MILP solver for the set partitioning formulation. They present results for various companies and propose a heuristic decomposition for larger problems. Anbil et al. \cite{Anbil1992} introduced a comprehensive method with a cost-minimization goal. In this research, presenting collaborative research between American Airlines Decision Technologies and IBM, numerous possible pairings (columns) are considered in a column pool, and a substantial amount is fed into the MILP solver. Then, several of the non-basic pairings are rejected, and pairings from the column pool are being inserted. The procedure is repeated until all pairings from the column pool have been taken into consideration \cite{Qiu2012}. This is only possible if the number of feasible pairings is not too large.
Because not all viable pairings are considered, the divergence from optimality can be significant. The same assessment can be made regarding all heuristic approaches.

To overcome the above limitations, more sophisticated approaches have been proposed over the years. To solve the CPP modelled as SPP, three time-horizons are generally studied: a daily, a weekly, and a monthly horizon~\cite{Brunner2013}, and the most prevalent resolution method since the 1990s is column generation inserted in branch-\&-bound \cite{Desaulniers1997,Barnhart1998}. The daily problem assumes that the flights are identical or relatively similar, for every day of the planning horizon, and that minimum cost pairings are generated for flights scheduled for a day. A daily cyclic solution is produced, where the number of crews present in every city in the evening is the same as in the morning. The weekly and monthly problems also assume repetitive schedules and function analogously except for the obvious longer time span that makes the master problem larger in row size. The daily solution is then unfolded over a typical week, such that one copy of each pairing is kept for each weekday, and infeasible copies are removed. Then, a cyclic weekly problem is solved, preserving as much as possible the unfolded daily solution. Similarly, the computed weekly solution is unfolded over the month, infeasible copies are removed, and the monthly problem is solved, preserving as much as possible the unfolded weekly solution.

For large fleets, it may take too long to globally solve the weekly problem or to globally re-optimize the monthly problem.
The rolling horizon approach is used to speed up the solution process \cite{Saddoune2013}. The horizon is divided into time slices of equal length (except maybe the last one), each one overlapping partially with the previous one. Then a solution is constructed greedily in chronological order by solving the problem restricted to each time slice sequentially, taking into account the solution of the previous slice, and the next one if it is a re-optimization, through additional constraints. When the size of the problem increases, the time slices need to be shorter to obtain problems that can be solved within a reasonable time period. When the time slices become shorter than the pairings, the quality of the solution is deeply impacted. Indeed, the pairings partially outside of the time slice cannot be moved to another base, and many deadheads are used to balance the workload between bases.
Recent studies have concentrated on weekly and monthly problems. Owing to the vacation period and differences in flight schedules, the monthly time horizon is the most accurate \cite{Brunner2013}.

\section{The improved algorithm for the crew pairing problem}\label{sec:Sol_Methods}

In this section, we present the structure of the \textit{Baseline} algorithm proposed by Desaulniers et al. \cite{desaulniers2019} in order to locate and explain the improvements introduced in this paper and to help understand the results on the performance of the algorithm.
First, Section \ref{sec:structure_alg} presents the structure of the algorithm in Figure \ref{fig:new_algo_spp} and focuses on the dynamic control strategies introduced to have a more stable algorithm producing better results. Then, Section \ref{sec:adapt_alg_ml} presents modifications to take advantage of clusters of flights with a high probability of being consecutive in a solution that may be different from the initial solution, as well as the modified complementary problem of IPS to reduce its density using the good information in the clusters. Finally, Section \ref{sec:imp_rolling_horizon} presents improvements to the rolling horizon algorithm for the monthly problem.

Before discussing the algorithmic improvements, we first shortly describe software improvements. Baseline works with GENCOL 4.5 using CPLEX 12.4. Commercial-GENCOL-DCA works with GENCOL 4.10 using CPLEX 12.6.3. Many adaptations were required because there was a three-year period between GENCOL 4.5 and GENCOL 4.10. During the experimentation of Commercial-GENCOL-DCA, we do not use parallel processors, as the emphasis was on improving the quality of the solution rather than reducing CPU time. This time will be reduced in the industry by using a workstation and parallel computing.

\subsection{The structure of the algorithm and improvements}\label{sec:structure_alg}

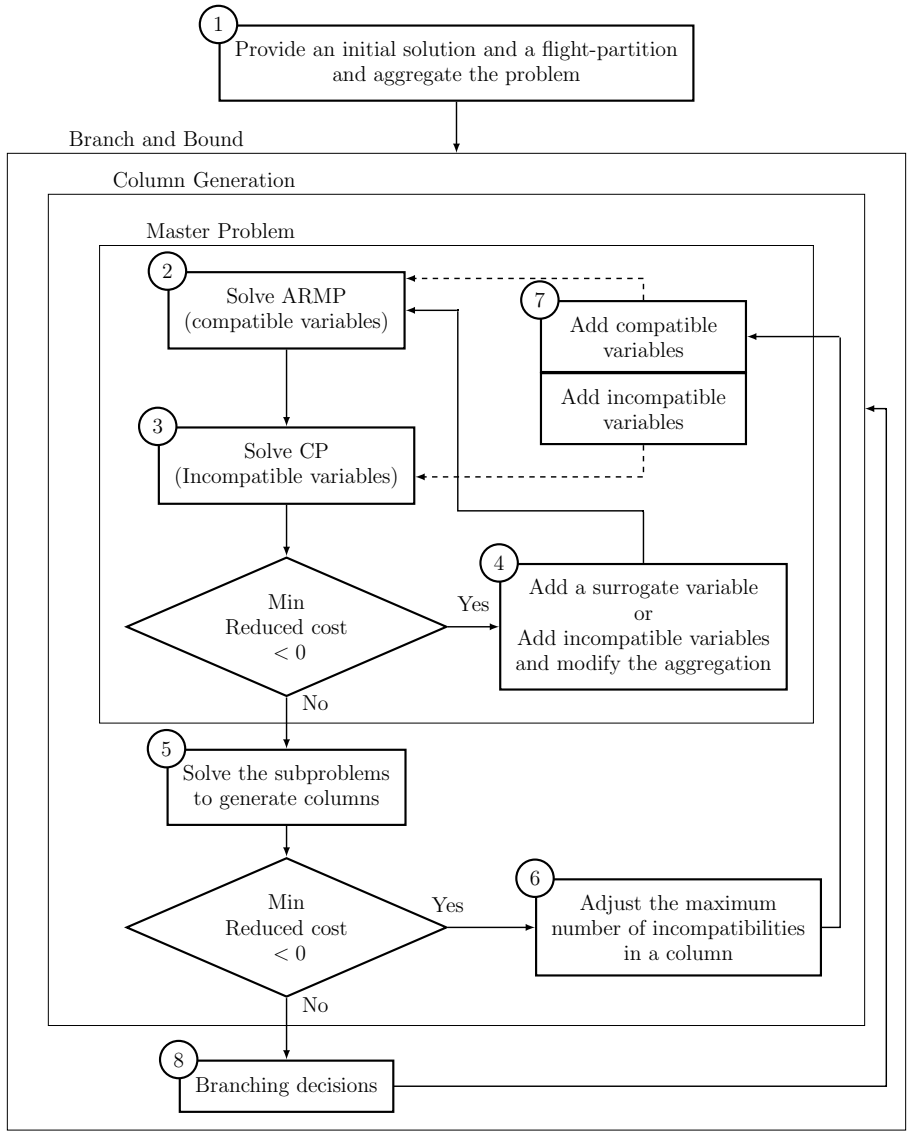
\begin{figure}
    \centering
    \resizebox{\linewidth}{!}{
    \large \begin{tikzpicture}[scale=1]
\tikzstyle{process} = [rectangle, minimum width=4cm, minimum height=1cm, inner sep=3mm, text centered,  very thick, align=center,draw=black]
\tikzstyle{decision} = [diamond, aspect=2.3, text centered,  very thick, align=center,  minimum width=1cm, minimum height=1cm, draw=black]
\tikzstyle{circmarker}=[circle,  text centered,  very thick, align=center,  minimum width=2mm,draw=black,fill=white]
\node (p2) [process] {Solve ARMP\\ (compatible variables)};
\node (c2)[circmarker] at (p2.north west) {2};

\node (p3) [process, below=1.5 cm of p2] {Solve CP\\ (Incompatible variables)};
\node (c3)[circmarker] at (p3.north west) {3};

\draw[thick, -latex] (p2.south)--(p3.north);
\node (d1) [decision, below=1 cm of p3] {Min\\ Reduced cost\\ ${}<0$};
\draw[thick, -latex] (p3.south)--(d1.north);

\node (p4) [process, right=1 cm of d1] {Add a surrogate variable \\ or \\ Add incompatible variables \\ and modify the aggregation};
\node (c4)[circmarker] at (p4.north west) {4};

\draw[thick, -latex] (d1.east)--(p4.west) node[midway, label=above:Yes]{};

\node (p6a) [process, above=2.3 cm of p4] {Add incompatible\\ variables};
\node (p6) [process, above=-0.02 cm of  p6a] {Add compatible\\ variables};
\node (c6)[circmarker] at (p6.north west) {7};

\draw[thick, dashed, -latex] (p6.north)-- (p6.north|-p2.15)--(p2.15);
\draw[thick, dashed, -latex] (p6a.south)-- (p6a.south|-p3.-5)--(p3.-5);

\node (p4n1) [rectangle=0.01mm, fill, outer sep=0mm,inner sep=0mm, above=1 cm of p4.north] {};
\node (p2e1) [rectangle=0.01mm,  fill, outer sep=0mm,inner sep=0mm,right=1 cm of p2.east] {};
\draw[thick, -latex] (p4.north)--(p4n1)--(p4n1-|p2e1)--(p2e1)--(p2.east);

\node[label={[xshift=-3cm,above left] {Master Problem}}, fit=(p2)(p3)(d1)(p4)(p6), draw, inner sep=5mm] {};

\node (p5) [process, below=1 cm of d1] {Solve the subproblems\\ to generate columns};
\node (c5)[circmarker] at (p5.north west) {5};

\draw[thick, -latex] (d1.south)--(p5.north) node[pos=0.1,label=right:No]{};

\node (d2) [decision, below=0.6 cm of p5] {Min\\ Reduced cost\\ ${}<0$};
\node (d2e1)[circle=0.1mm, red, outer sep=0mm,inner sep=0mm,below=-1 cm of d2] {};

\draw[thick, -latex] (p5.south)--(d2.north);

\node[label={[xshift=-3cm,above left] {Column Generation}}, fit=(p2)(p3)(d1)(p4)(p6)(p5)(d2e1), draw, inner sep=15mm] {};

\node (p6e1)[rectangle=0.1mm, red, outer sep=0mm,inner sep=0mm,right=1.8 cm of p6.east] {};

\node (p8) [process, right=1.7 cm of d2] {Adjust the maximum \\ number of incompatibilities \\ in a column};
\node (c8)[circmarker] at (p8.north west) {6};

\draw[thick, -latex] (d2.east)--node [pos=0.1,label=above:Yes]{}(d2.east)--(p8);
\draw[thick, -latex] (p8)--(p8-|p6e1)--(p6e1)--(p6.east) ;

\node (p7) [process, below=1.2 cm of d2] {Branching decisions};
\node (p7e1)[rectangle=0.1mm, red, outer sep=0mm,inner sep=0mm,below=-2 cm of p7] {};

\node (c7)[circmarker] at (p7.north west) {8};

\draw[thick, -latex] (d2.south)--(p7.north) node[pos=0.1,label=right:No]{};

\node(BAB)[label={[xshift=-4cm,above left] {Branch and Bound}}, fit=(p2)(p3)(d1)(p4)(p6)(p5)(d2)(p7e1), draw, inner sep=23mm] {};

\node (p6e2)[rectangle=0.1mm, fill,  outer sep=0mm,inner sep=0mm,right=2.7cm of p6a.east] {};

\node (p6e3)[rectangle=0.1mm,  outer sep=0mm,inner sep=0mm,left=4 mm of p6e2] {};

\draw[thick, -latex,line cap=rect, ] (p7.east)--(p7.east-|p6e2)--(p6e2)--(p6e3) ;

\node (p1) [process, above=1 cm of BAB] {Provide an initial solution and a flight-partition \\ and aggregate the problem};
\node (c1)[circmarker] at (p1.north west) {1};

\draw[thick, -latex] (p1.south)--(BAB.north);
\end{tikzpicture}}
    \caption{The improved algorithm for the SPP type}
    \label{fig:new_algo_spp}
\end{figure}

As shown in Figure~\ref{fig:new_algo_spp}, in box 1, the initial solution is a set of feasible pairings. These pairings do not necessarily cover all flights. It is better to have a partial solution than to start from scratch. This solution permits a warm-start of CPLEX with few artificial variables. The flight-partition is a set of clusters of flights. Within each cluster, flights have a high probability of being done consecutively by the same crew. Unlike Baseline, where clusters are the pairings of the initial solution, in Commercial-GENCOL-DCA, the clusters may be different from the pairings of the initial solution, allowing the use of clusters with fewer mispredicted flight-connections. Indeed, a heuristic solution contains some mispredicted connections to include all flights and still be feasible. It permits to use clusters produced by ML containing only sequences of flights with a high probability of staying together. Note that we are not aware of a ML approach that can directly construct a feasible solution (taking into account the many complex airline-dependent costs and constraints that are not necessarily available to the ML system at training time). The construction of clusters by ML will be explained in Section~\ref{sec:ML_Model}. The mathematical programming part of the algorithm is a Branch-\&-Bound using column generation at each node of the branching tree.

In box 2, the Aggregated Reduced Master Problem (ARMP) contains only the compatible pairing variables generated up to date in the column generation process. Recall that a variable is said to be compatible with respect to the partition if the flights covered by the pairing correspond to a concatenation of some clusters. Otherwise, this variable is declared incompatible. Observe that the covering constraints of the flights in a cluster are identical for the compatible variables of ARMP, which justifies keeping only the covering constraints of the first flight in each cluster in the ARMP. The ARMP also contains a few supplementary linear constraints; for example, the base constraints limiting the flight time per base. The ARMP, with only one constraint per cluster and only the compatible variables, permits to rapidly improve the solution. This smaller non-degenerate problem or with very few degeneracy is solved efficiently with the primal simplex.

In box 3, the complementary problem (CP) contains only the incompatible pairing variables generated up to date. This linear program finds a compatible convex combination of incompatible columns, with minimal reduced cost. Let $S$ be the set of non-zero variables in this solution. These variables have the same value. Let $A_S$ be the surrogate column obtained by the summation of the columns in $S$. The dual solution of CP is combined with the dual solution of the ARMP to obtain dual variables for all flights in the problem. Section \ref{sec:adapt_alg_ml} gives more details on the improvements added to solve the CP more efficiently.

In box 4, we select between two options. The first is to add the surrogate column in the ARMP. This compatible column can be added without changing the aggregation and the number of constraints in the ARMP. When the ARMP is non-degenerate, the cost of the solution decreases at the first pivot. The ARMP of the CPP with few additional constraints has few degeneracy, and the solution is improved most of the time with one or very few pivots. The second is to add in the ARMP, the columns of the variables in the set $S$. Some clusters of the partition are broken to make compatible the added variables. The incompatible variables becoming compatible are also added to the ARMP. When the ARMP in non-degenerate, the first $|S|-1$ pivots are degenerate, and the $|S|^{th}$ pivot improves the solution.

We develop the following strategy. In the first iterations of the column generation, we use the first option. The cost decreases rapidly with this small ARMP. When the tailing effect appears, we observe that some surrogate variables enter and leave the basis with small step sizes in the simplex pivots. This is due to the fact that minimization on many terms of a dense column in the exit criteria of the simplex has smaller values and creates pivots with small steps.

Indeed, a simplex tableau with few non zeros in it is said to have a low density, and be sparse. Informally, a problem with a sparse initial tableau may be referred to as a sparse problem.
Singleton columns (containing a single entry) do not increase the density of the tableau when they are pivoted on. Similarly, doubleton columns (containing two entries) can only increase the density slightly when they are pivoted on. Singleton and doubleton columns are called sparse columns. The density of the simplex tableau affects the density of the pivot column and row. As pivoting uses the pivot column and row across the tableau, the density of the pivot row and column affects the amount of work required. It is therefore desirable to preserve the sparsity of the tableau during the solution process.

In the situation where the tailing effect appears, we replace the surrogate variables that take a positive value in the solution of ARMP by the columns composing these combinations (which may require adjusting the current partition) while the other surrogate variables are discarded from the ARMP. The optimization continues with the second option. The cost function of this less dense problem restarts to decrease more rapidly. Furthermore, the solution becomes less fractional with this less dense problem. Replacing the surrogate variables by the original variables before branching permits only branching on original variables.

In box 5, when the solution of CP has null reduced cost, the sub-problem needs to be solved to generate columns with negative reduced cost, if possible. To be more efficient, we introduce the following dynamic strategy: we go to the sub-problem when the reduced cost is above a negative threshold. Instead of adding existing incompatible columns with a small potential of solution improvement, we give priority to generating columns with a better potential for improvement. At each column generation iteration, this threshold is increased up to zero to reach the optimality criteria. There is a sub-problem per crew base and per starting day for pairings. Each sub-problem has a shortest path problem structure with resource constraints modeling the rules, ensuring that the paths are feasible pairings.

In box 6, we control the partial pricing strategy of MPDCA (Elhallaoui et al. \cite{Elhallaoui2008_multi}). This strategy uses the degree of incompatibility of a column, which is the number of times an incompatible column enters or exits in the middle of a cluster. This value can be computed in the sub-problem when a column is generated. MPDCA proceeds through a predetermined sequence of phases, typically, phases $k$ $=$ $0$, $1$, $2$, $\dots$ In phase $k$, only pairings with a degree of incompatibility not exceeding $k$ can be generated by the pricing problems. To impose this constraint in the pricing problem, an additional incompatibility degree resource is considered. In \textit{Baseline}, the value of $k$ is increased of one unit, from $0$ to $2$, when the objective improvement becomes too small. More explanation can be found in Desaulniers et al. \cite{desaulniers2019}. In Commercial-GENCOL-DCA, a more dynamic strategy manages the value of $k$ using $T_1$, a threshold on $T$, the min reduced cost in the sub-problem, and $M$ a maximum on $N$ the number of fractional variables in the solution of ARMP. We used $M = p~\times$ number of flights, with $p = .3$ and $p = .6$.
The rule is the following:
\begin{enumerate}
    \item if $T > T_1$ and $N < M$, then increase $k$ of one unit up to $3$;
    \item if $T > T_1$ and $N > M$, replace $T_1$ by \mbox{\large$\frac{T_1}{2}$};
    \item if we reach $T > T_1$ and $N > M$ and $k =3$, we take few branching decisions on variables very close to $1$. It reduces the number of fractional variables and we take $k = 1$ and come back to rule $1$. This rule maintains $k$ as small as possible, therefore gives priority to less dense columns and produces less fractional solutions in ARMP by combining less dense columns. This point will be explained in Section \ref{sec:adapt_alg_ml}.
\end{enumerate}

In box 7, the compatible and incompatible variables are identified with a good data structure according to the definition of compatibility given in box 2. This easily applies to CPPs, since flights can be ordered by departure time. A path corresponding to a variable is stored as a sequence of pointers to the next flight using the departure-time ordering, and clusters are stored in a similar manner. In addition, in clusters, each flight in a path has a pointer to itself. Starting at the beginning of the path of a variable, it is easy to check if the path is the concatenation of clusters, and if it is compatible. Compatible variables are added to the ARMP. Incompatible ones are added to the CP. After adding new variables, the ARMP is resolved first without changing the partition. The ARMP has priority as long as its least reduced cost variable is less than the least reduced-cost of incompatible variables multiplied by a predetermined multiplier (smaller than $1$).

In box 8, we use a partial exploration of the branching tree.
Since column generation solves the sub-problem at optimality and has access to a very large number of columns, the LP relaxation provides a very good lower bound. Furthermore, MPDCA provides a primal solution with a small number of fractional variables.
The lower bound and the primal solution give useful information to efficiently explore the branching tree. We first use the column fixing algorithm as follows (for instance, see \cite{Saddoune2013}): At each node of the search tree, several columns taking a fractional-value in the current ARMP solution are selected, and their values are set to one. These columns are selected in decreasing order of their values because columns with larger values, in general, increase less the value of the lower bound when they are set to one than columns with smaller values. After imposing a pairing $p$, we remove from the ARMP and CP all columns containing the flights covered by $p$, and from the sub-problem networks all arcs representing those flights. These decisions reduce rapidly the size of the problems to solve. When there are no variables large enough to fix, we use the arc fixing strategy. We fix to one the arcs with a large value. Unfortunately, such diving heuristic may sometimes make bad decisions and produce a poor integer solution. To reduce the weakness of the diving branching, we use the Retrospective Branching \cite{quesnel2017}. This algorithm detects and revises, without backtracking, poor decisions made previously in the search tree. These decisions are selected from a list of risky decisions that is maintained during the branching process. A risky decision is a column, or an arc fixed even if its value was smaller than a threshold. When the relative gap $q_{i}$ between the values $z_{i}$ and $z_{0}$ of the solutions computed at a node $i$ in the search tree and at the root node (i.e., $q_{i}$ = \mbox{\Large$\frac{z_{i}-z_{0}}{z_{0}}$}) exceeds a relative estimate gap for a good integer solution, the algorithm ejects risky decisions from the current solution without backtracking. This ejection is performed by adding a constraint on the number of risky decisions in the solution. The Retrospective Branching improves the integer solution by 25\% of the optimality gap compared to the diving branching.

\subsection{The adaptation of the algorithm to use Machine Learning initial information}\label{sec:adapt_alg_ml}

IPS and Baseline use a basis $B$ compatible with the initial solution, to compute $\bar{A} = B^{-1}A$ in the construction of CP. This computation can be time-consuming for large problems. To reduce the density of $A$, we use a transformation matrix $M$ obtained from the information in the clusters. Note that multiplying by $M$ does not change the set of feasible solutions, since we will construct $M$ to be injective. Let us then consider that the flight covering constraints are in the order of flight start time. The transformation matrix $M$ is block-diagonal, with a block for each cluster. Figure \ref{fig:figure5} presents the structure of $M$ and the block structure associated with the clusters including the flights $\ell_1,\dots,\ell_k$.

\begin{figure}[H]
\[\resizebox{\linewidth}{!}{$
M=\begin{pmatrix}
M^1 &&&\multicolumn{2}{c}{\raisebox{-2ex}[0ex][0ex]{\scalebox{1.5}{$0$}}}\\
& M^2 &&&\\
&& \ddots &&\\
\multicolumn{2}{l}{\hspace{4mm}\raisebox{1ex}[0ex][0ex]{\scalebox{1.5}{$0$}}}&&& M^{|L|}
\end{pmatrix}
\text{ with }
\underset{\substack{ \ell_1 \leq i \leq \ell_k \\ \ell_1 \leq j \leq \ell_k}}{\displaystyle M^\ell}\hspace{-1mm}=
\begin{pmatrix}
1 &&&&\multicolumn{2}{c}{\raisebox{-3ex}[0ex][0ex]{\scalebox{1.75}{$0$}}}\\
-1 & \phantom{-}1 &&&&\\
& -1 & \phantom{-}1 &&&\\
&&& \ddots &&\\
&&& -1 & \phantom{-}1 &&\\ 
\multicolumn{2}{l}{\hspace{6mm}\raisebox{2ex}[0ex][0ex]{\scalebox{1.75}{$0$}}}&&& -1 & \phantom{-}1\\
\end{pmatrix}
$}
\]
\caption{The transformation matrix $M$ and the block for one cluster}\label{fig:figure5}
\end{figure}

Computing $A^1 = M . A$ consists in modifying $A$ by subtracting the row $i$ from the row $i+1$, for $i=1 \dots m$. Figure \ref{fig:product} presents an example of $M . A$ for a sample of typical columns of $A$. Consider $2$ clusters of flights $(1, 2, 3); (4, 5, 6)$. The sample of columns is:

\begin{table}[H]
\begin{tabular}{cll}
Column & Path               & Property                                          \\
1      & (1, 2, 3)          & compatible, covers cluster 1                      \\
2      & (1, 2, 3, 4, 5, 6) & compatible, covers clusters 1 and 2               \\
3      & (1, 2, 4, 5, 6)    & incompatible, exits from cluster 1 after flight 2 \\
4      & (2, 3)             & incompatible, enters in cluster 1 before flight 2 \\
5      & (2, 3, 4, 5)       & incompatible, enters in cluster 1 before flight 2 \\
       &                    & and exits from cluster 2 after flight 5            \\
\end{tabular}
\end{table}

\begin{figure}[H]
\[
\resizebox{\linewidth}{!}{$
{\setlength{\arraycolsep}{3pt}
\begin{array}[t]{r|*{3}{c}|*{3}{c}|}
\multicolumn{1}{l}{} &\multicolumn{6}{c}{$M$}\\\multicolumn{7}{c}{}\\\cline{2-7}
1&1&&&&&\\
2&-1&\phantom{-}1&&&\raisebox{-0.5ex}[0ex][0ex]{\scalebox{1.3}{$0$}}&\\
3&&-1&\phantom{-}1&&&\\\cline{2-7}
4&&&&1&&\\
5&&\raisebox{-0.5ex}[0ex][0ex]{\scalebox{1.3}{$0$}}&&-1&\phantom{-}1&\\
6&&&&&-1&\phantom{-}1\\\cline{2-7}
\end{array}}\quad \raisebox{-2.76cm}[0ex][0ex]{\scalebox{1.5}{$\cdot$}}\enskip
\begin{array}[t]{*{5}{c}}
\multicolumn{5}{c}{A}\\
1&2&3&4&5\\\hline
1&1&1&0&0\\
1&1&1&1&1\\
1&1&0&1&1\\
0&1&1&0&1\\
0&1&1&0&1\\
0&1&1&0&0\\[-8pt]
\multicolumn{2}{c}{{\underbrace {\hspace{9mm}}_{\tikzmark{comp}}}}&\multicolumn{3}{c}{{\underbrace {\hspace{14mm}}_{\tikzmark{incomp}}}}
\end{array}\quad \raisebox{-2.75cm}[0ex][0ex]{\scalebox{1.1}{$=$}}\quad
\begin{array}[t]{r@{\hspace{5.1mm}}r@{\hspace{2.75mm}}r@{\hspace{5.1mm}}r@{\hspace{2.75mm}}r}
\multicolumn{5}{c}{A^1}\\
1&2&3&4&5\\\hline
1&1&1&0&0\\
0&0&0&1&1\\
0&0&-1&0&0\\
0&1&1&0&1\\
0&0&0&0&0\\
0&0&0&0&-1\\
\end{array}
$}
\begin{tikzpicture}[overlay,remember picture]
    \node[anchor=north]
    at ( $ (pic cs:comp) +(-2mm,2mm) $ )
    {\scriptsize compatible};
    \node[anchor=north]
    at ( $ (pic cs:incomp) +(0mm,2mm) $ )
    {\scriptsize incompatible};    
\end{tikzpicture}
\]
\caption{Example of computing the matrix $A^1$}\label{fig:product}
\end{figure}

The matrix $A^1$ has the following properties:
\begin{enumerate}
\itemsep-0.25mm
    \item a compatible column has a $1$ on the row of the first flight of each cluster covered by this column and a $0$ for the other flights (columns $1$ and $2$);
    \item an incompatible column leaving a cluster just before the flight $i$ has $-1$ on the row of flight $i$ (columns $3$ and $5$);
    \item an incompatible column entering in a cluster just before the flight $i$ has $+1$ on the row of flight $i$ (columns $4$ and $5$);
    \item an incompatible column covering the first flights of a cluster has $+1$ on the row of the flight.
\end{enumerate}

We reorganize $A^1$ by moving in the upper part the rows of the first flights of each cluster and, in the lower part, the remaining rows. We obtain the following matrix 
$\begin{pmatrix}
A_{C}^{1} & A_{I}^{1} \\
A_{C}^{2} & A_{I}^{2} \\
\end{pmatrix}\rule[-6.5mm]{0pt}{14.5mm}$; where $A_{C}^{1}$ ($A_{C}^{2}$) are the upper (lower) part of the compatible columns, and $A_{I}^{1}$ ($A_{I}^{2}$) are the upper (lower) part of the incompatible columns.

In what follows, we detail a few additional remarks:
\begin{itemize}
\itemsep-0.25mm
    \item $A_{C}^{1}$ is the matrix of the ARMP;
    \item $A_{C}^{2}$ = 0, it justifies removing these rows in the ARMP;
    \item $M . A_j = A_j^2 = 0$ is an easy criteria to identify if column $j$ is compatible;
    \item the number of non-null elements in a column of $A^2_j, j \in I$ is the number of incompatibilities in this column;
    \item $A^2_I$ has low density using the strategy described in box 6, giving priority to the generation of columns with few incompatibilities.
\end{itemize}

The complementary problem CP finds a combination of compatible columns with a minimum reduced cost. After the matrix transformation, the constraints of CP can be rewritten as:
\begin{align}
A_I^2 v &= 0 \label{Eq:CP1} \\
w . v &= 1 \label{Eq:CP2} \\
v &\geq 0
\end{align}

Asking $A_I . v$ to be compatible is equivalent to asking $A_I^2 . v = 0$. Without the constraint (\ref{Eq:CP2}), where $w = (1, \cdots, 1)^\top$ is of dimension dictated by the context, the problem is unbounded and produces an extreme ray. With this constraint, the problem produces an extreme point. The columns of $A_I^2 v$ are obtained easily by transforming the columns generated using the data structures described in Box 6.

IPS and Baseline use a basis $B$ compatible with the initial solution, to compute $\bar{A} = B^{-1}A$ in the construction of CP. There are many bases $B$ compatible with the solution for a degenerate problem. We can select a basis that is easier to invert by replacing the null variables in the basis by high-cost artificial variables. The original zero variables become non-basic. This permits to replace a part of the basis by an identity matrix (Zaghrouti et al. \cite{zaghrouti2018}). $B_1$, this new basic matrix, is used by \textit{Baseline} and Commercial-GENCOL-DCA. We use $B_1$ to speed up the ARMP. For CP, there is still a difficult part in the matrix $B_1$, and the computation of $\bar{A} = B^{-1}A$ can be expensive. In Commercial-GENCOL-DCA, we use the transformation matrix $M$ to reduce the density of $A$. Indeed, this problem with fewer constraints and a low-density matrix can be solved more easily. Because CP is severely degenerate, primal simplex is not a good algorithm to solve it. Instead, the dual simplex is a more appropriate algorithm. The dual is likely not degenerate because the reduced costs in the objective are real numbers, such that the probability of having two subsets of variables with the same cost is very low. Furthermore, the use of the less dense matrix $A^1$ is an adaptation of the algorithm to take advantage of the good information provided by ML. Indeed, the clusters provided by ML contain more reliable information than the initial solution produced with a heuristic. While the heuristic makes a set of compromises to satisfy the constraints, ML, using the abstention strategy presented in Section \ref{sec:Cluster_Construction}, keeps only the good information and discards what is less reliable.

\subsection{Improvements of the rolling horizon algorithm for the monthly problem}\label{sec:imp_rolling_horizon}

We improve the integration in a rolling horizon approach, in order to use clusters tailored for the flights of each window and connecting well with the schedule of the previous window. In order to present our contributions, we first present the optimization process used in GENCOL init and Baseline:

\paragraph{GENCOL init} Consider a standard monthly solution called ``GENCOL init'', obtained with the GENCOL solver (without DCA). In this approach, the problem is solved by a ``rolling horizon'' approach. Because the GENCOL solver (without DCA) is able to solve up to a few thousand flights per window, it is constrained to use two-day windows and a one-day overlap period. This means that the month is divided into overlapping time slices of equal length. Then a solution is constructed greedily in chronological order by solving the problem \emph{restricted} to each time slice sequentially, taking into account the solution of the previous slice through additional constraints.

\paragraph{Baseline}
The constraint aggregation approach is used in a rolling-horizon procedure (seven-day windows and a two-day overlap period) for the monthly CPP.
\textit{Baseline} is fed with the pairings of the solution ``GENCOL init'' as an initial solution, thus obtaining a solution that we consider as a baseline for comparison.
These pairings are used as initial clusters. These clusters are inadequate, since (1) ``GENCOL init'' uses narrow windows and makes many compromises to produce a feasible solution, and (2) the initial pairings in the next seven-day windows to optimize are in conflict with the new pairings in the overlapping days, generated by the optimizer in the previous window.

\paragraph{Commercial-GENCOL-DCA}
For each experiment, we use ML to construct an initial DCA partition and feed it to Commercial-GENCOL-DCA as initial clusters to solve the monthly CPP. In Commercial-GENCOL-DCA the clusters are dynamically generated by ML after the solution of the previous windows, which permits to have clusters connecting well with the pairings starting in the previous five days.

\section{Machine Learning model}\label{sec:ML_Model}

This section describes the ML predictor constructed to provide the probabilities of flights being performed consecutively by the same crew. The prediction problem is first stated and formulated in Section \ref{sec:Prediction_Problem_Formulation}. The proposed prediction model is disclosed in Section \ref{sec:NN_architecture}. Section \ref{sec:_Transf_Input} presents the input transformations needed to get an effective prediction model. Upon the finalization of the model training, and in order to construct monthly pairings, we use four heuristics presented in Section \ref{sec:Cluster_Construction}.

\subsection{Prediction problem formulation}\label{sec:Prediction_Problem_Formulation}

In our solution method for the overall CPP, we provide the optimizer with an initial partition of flights into clusters. Each cluster represents a sequence of flights with a high probability of being consecutive in the same pairing in the solution. To construct each cluster, we need the following:

\begin{itemize}
\itemsep-0.25mm
    \item Information on where and when a crew begins a pairing. This information makes it possible to identify whether a flight is the beginning of a pairing;
    \item For each incoming flight to a connecting city, predict what the crew is going to do next: layover, flight, or end of a pairing. If it is the second case (flight), then we further predict which flight the crew will undertake.
\end{itemize}

Since the end of a pairing depends on the maximum number of days in a pairing permitted by the airline company, we will solely rely on this number as a hyperparameter. In other words, there is no need to predict the end of the pairing. Therefore, with regards to the pairing construction, we propose to decompose the second task into two sub-tasks. The first is predicting whether the crew will make a layover; the second is predicting the next flight, under the assumption that the crew will always take another flight.

\begin{figure}[H]
    \centering
    \resizebox{0.7\linewidth}{!}{
    \large \begin{tikzpicture}
\definecolor{lightblue}{HTML}{aed8e5}
\definecolor{dodgerblue}{HTML}{1b8ff9}
\definecolor{mypink}{HTML}{fcc2cd}
\tikzstyle{ellipsenode} = [ellipse, draw, minimum width=2cm, minimum height=1.2cm, align=center, very thick, fill=lightblue]
\tikzstyle{circlenode} = [circle, draw, minimum width=1cm, minimum height=1cm, align=center, very thick, fill=dodgerblue]
\tikzstyle{squarenode} = [rectangle, draw, minimum width=0.7cm, minimum height=0.7cm, align=center, very thick, fill=mypink]
\node[ellipsenode] (A) {Base};
\node[circlenode, right=0.65cm of A] (B1) {};
\node[circlenode, right=0.2cm of B1] (B2) {};
\node[circlenode, right=0.5cm of B2] (B3) {};
\node[squarenode, right=0.8cm of B3] (C) {};
\node[circlenode, right=0.8cm of C] (D1) {};
\node[circlenode, right=0.4cm of D1] (D2) {};
\node[ellipsenode, right=0.65cm of D2] (E) {Base};
\coordinate (L1) at ($(B1.north west)+(-0.25,0.25)$);
\coordinate (L2) at ($(B3.south east)+(0.25,-0.25)$);
\coordinate (L3) at ($(C.north west)+(-0.235,0.235)$);
\coordinate (L4) at ($(C.south east)+(0.235,-0.235)$);
\coordinate (L5) at ($(D1.north west)+(-0.25,0.25)$);
\coordinate (L6) at ($(D2.south east)+(0.25,-0.25)$);
\draw[densely dashed] (L1) rectangle (L2)node[midway,yshift=-0.9cm]{Duty period};
\draw[densely dashed] (L3) rectangle (L4)node[midway,yshift=-0.9cm]{Layover};
\draw[densely dashed] (L5) rectangle (L6)node[midway,yshift=-0.9cm]{Duty period};
\draw[very thick](A.east)--(B1.west);
\draw[very thick, red] (B1.east)--(B2.west);
\draw[very thick, red] (B2.east)--(B3.west);
\draw[very thick, red] (B3.east)--(D1.west);
\draw[very thick, red] (D1.east)--(D2.west);
\draw[very thick, -stealth, red] (D2.east)--(E.west);
\node[circlenode, below left=1cm and 2cm of E, minimum width=0.5cm, minimum height=0.5cm, text width=0cm] (F) {};
\node[squarenode, below=0.2cm of F,minimum width=0.5cm, minimum height=0.5cm, text width=0cm] (G) {};
\node[right, xshift=0.3cm] at (F) {Operated flight};
\node[right, xshift=0.3cm] at (G) {Rest};
\draw[red, ultra thick] ($(G.west)+(0,-0.8)$) -- ($(G.east)+(0,-0.8)$)node[right, xshift=0.02cm, black]{Flight-Connection};
\end{tikzpicture}}
    \caption{Illustration of a crew pairing. Here the flight-connection variable (in red) is defined to determine the next flight that a crew is going to perform, given an incoming flight}
    \label{Fig:CP_Illustration}
\end{figure}
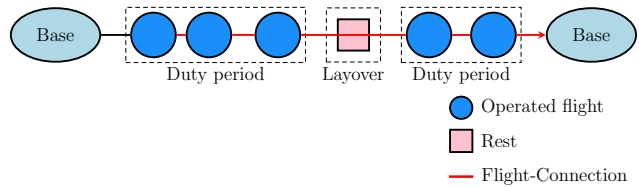

On the one hand, layovers are highly correlated with the duration of the connection, although there is no fixed threshold for the number of hours a crew must stay in an airport to make a layover. On the other hand, predicting the next flight is a much more complicated prediction problem. We call this the \textbf{flight-connection problem}, which is the focus of our ML approach described in the next sections (see Figure \ref{Fig:CP_Illustration}).

Similar to Yaakoubi et al. \cite{Yaakoubi2019}, we transform the data to build a flight-connection prediction problem (a supervised ML multiclass classification problem) where the goal is to predict the next flight that an airline crew should follow in their schedule given the previous flight. The classification problem is thus the following: given the information about an incoming flight in a specific connecting city, choose among all the possible departing flights from this city (which can be restricted to the next 48 hours) the one that the crew should follow. These departing flights will be identified by their flight code and departing city, and day (about 2 000 possible flight codes in our dataset). Note that different flights may share the same flight codes in some airline companies, as flights performed multiple times a week usually use the same flight code. Nevertheless, a flight is uniquely identified by its flight code, departing city, and day.

Each flight can be described by the following five features that we can use in our classification algorithm:
\begin{itemize}
\itemsep-0.25mm
    \item City of origin and city of destination ($\sim$200 categories);
    \item Aircraft type (5 categories);
    \item Duration of flight (in minutes)
    \item Time (with date) of arrival (for an incoming flight) or of departure (for a departing flight).
\end{itemize}

\subsection{Neural network architecture}\label{sec:NN_architecture}

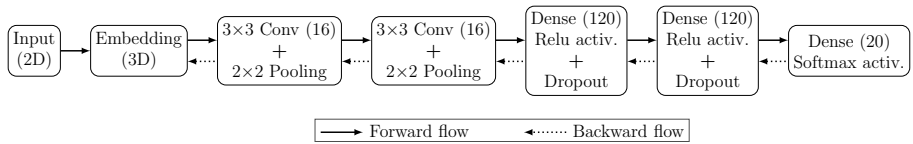
\begin{figure}
\vspace{-0.2cm}
{\centering
\resizebox{\textwidth}{!}{ 
\begin{tikzpicture}[font=\large]
\node[align=center,draw,anchor=west, rounded corners=2mm, minimum width=1cm, minimum height=1.2cm] (A) at (0,0) {Input \\ (2D)};
\node[align=center,draw,anchor=west, rounded corners=2mm, minimum width=1cm, minimum height=1.2cm] (B) at ([xshift=7mm]A.east) {Embedding \\ (3D)};
\node[align=center,draw,anchor=west, rounded corners=2mm, minimum width=1cm, minimum height=1.2cm] (C) at ([xshift=7mm]B.east) {3$\times$3 Conv (16) \\ \textbf{+} \\ 2$\times$2 Pooling};
\node[align=center,draw,anchor=west, rounded corners=2mm, minimum width=1cm, minimum height=1.2cm] (D) at ([xshift=7mm]C.east) {3$\times$3 Conv (16) \\ \textbf{+} \\ 2$\times$2 Pooling};
\node[align=center,draw,anchor=west, rounded corners=2mm, minimum width=1cm, minimum height=1.2cm] (E) at ([xshift=7mm]D.east) {Dense (120) \\ Relu activ. \\ \textbf{+} \\ Dropout};
\node[align=center,draw,anchor=west, rounded corners=2mm, minimum width=1cm, minimum height=1.2cm] (F) at ([xshift=7mm]E.east) {Dense (120) \\ Relu activ. \\ \textbf{+} \\ Dropout};
\node[align=center,draw,anchor=west, rounded corners=2mm, minimum width=1cm, minimum height=1.2cm] (G) at ([xshift=7mm]F.east) {Dense (20) \\ Softmax activ.};

\draw[line width=1pt,-latex] (A.east)--(B.west);

\draw[line width=1pt,-latex] ([yshift=2.5mm]B.east)--([yshift=2.5mm]C.west);
\draw[line width=1pt,latex-, dotted] ([yshift=-2.5mm]B.east)--([yshift=-2.5mm]C.west);

\draw[line width=1pt,-latex] ([yshift=2.5mm]C.east)--([yshift=2.5mm]D.west);
\draw[line width=1pt,latex-, dotted] ([yshift=-2.5mm]C.east)--([yshift=-2.5mm]D.west);

\draw[line width=1pt,-latex] ([yshift=2.5mm]D.east)--([yshift=2.5mm]E.west);
\draw[line width=1pt,latex-, dotted] ([yshift=-2.5mm]D.east)--([yshift=-2.5mm]E.west);

\draw[line width=1pt,-latex] ([yshift=2.5mm]E.east)--([yshift=2.5mm]F.west);
\draw[line width=1pt,latex-, dotted] ([yshift=-2.5mm]E.east)--([yshift=-2.5mm]F.west);

\draw[line width=1pt,-latex] ([yshift=2.5mm]F.east)--([yshift=2.5mm]G.west);
\draw[line width=1pt,latex-, dotted] ([yshift=-2.5mm]F.east)--([yshift=-2.5mm]G.west);

\node[line width=0.5pt,draw=black!70,anchor=north west] at ([xshift=3cm,yshift=-10mm]B.south east) {\tikz[baseline=-1mm]\draw[line width=1pt,-latex]  (0pt,0pt) -- (10mm,0pt);  Forward flow \hspace{1cm} \tikz[baseline=-1mm]\draw[line width=1pt,-latex,dotted]  (10mm,0pt) -- (0pt,0pt);  Backward flow};
\end{tikzpicture}
}}
\caption{The architecture diagram for the NN predictor. Note that in this example, we use 2 convolutional layers, 3 $\times$ 3 as filter size for the convolutional layers, 2 $\times$ 2 for pooling, 2 hidden layers with 120 neurons in each. These values do not necessarily represent the best configuration of hyperparameters, which will be set using Bayesian Optimization.}
\label{fig:flowchart}
\end{figure}

As shown in Yaakoubi et al. \cite{Yaakoubi2019,yaakoubi2019these}, Neural Networks (NN) have shown great potential to extract meaningful features from complex inputs and obtain high accuracy. Therefore, we use a modified version of their NN predictor, as shown in Figure~\ref{fig:flowchart}.

First, using standard encoding, the city code, and aircraft type features are treated as numeric values. This means that cities with close values are treated similarly by the NN even though the codes are somewhat arbitrary. A more meaningful encoding for such categorical features is to use one-hot encoding. By fully connecting each one-hot encoding of a categorical feature to a separate hidden layer, we get an embedding layer of dimension $d$ for this feature ($d$ is a hyperparameter). The $C \times d$ parameter matrix (where $C$ is the number of possible categorical values) represents the $d$-dimensional encoding for each of the $C$ values, and this encoding is learned during the NN training. The embedding layer approach thus learns a $d$-dimensional representation of each city, and one could explore which city is similar to another one in this space. By concatenating the embedding layer for each categorical feature with the other numeric features (such as hours and minutes), we get an $n_d$-vector, where $n_{d}$ is the dimensionality of the representation of one flight that is fed into the NN.

Second, we use a multi-layer CNN. CNNs are a particular type of NNs for processing grid-type data, e.g., time series (1D), image data (2D), and medical CT or MRI scans (3D). The convolution layer convolves the input matrix with a filter (kernel) of predefined size. The kernel weights are updated during the learning process. Our convolutional layer consists of successive layers organized hierarchically; each layer has convolutions with learned filters, followed by point nonlinearity and a sub-sampling operation called pooling. Successive convolution layers automatically extract the most relevant features from the input data and insert them into the final classification layers of the network. Note that if we use a large number of hidden layers and neurons per layer, NNs gives similar results to CNNs in our case. But, using CNNs gives more stable results, is far less time consuming, and takes into consideration the input format. More intuition behind using CNNs is explained in Section \ref{sec:_Transf_Input}. The output layer is a dense layer with \texttt{softmax} as the activation function, defined by Eq. \ref{Eq:Softmax}, where $x$ is a vector of dimension $K$ given by the last hidden layer. We use categorical cross-entropy as a loss function and standard categorical accuracy on the test set as a performance metric.

\begin{equation}
    \operatorname{softmax}(x)_i = \frac{e^{x_i}}{\sum_{k = 1}^{K}e^{x_k}}, \ \forall \, i \in \llbracket 1~,~20  \rrbracket
    \label{Eq:Softmax}
\end{equation}

Third, we use dropout~\cite{srivastava14a} to prevent overfitting in NNs.
It is a generalization method and consists of applying (element-wise product) a binary mask to each layer outputs during training and cached for future use on back-propagation. Since each neuron is susceptible to being masked, neurons learn the ability to adapt to the lack of information, thus making the NN more robust and avoid overfitting. We also use batch normalization~\cite{batchnorm} between the activation function and the dropout, to normalize the activations over the current batch in each hidden layer. Indeed, the layers adjust their parameters during back-propagation with the assumption that their input distribution stays the same. But, all the layers are updated with back-propagation, thereby changing each of their outputs. This changes the inputs of the layers as each layer gets its inputs from the previous layer. This is called \textit{internal covariate shift}. Hence, the layers experience a constantly changing input distribution. Batch normalization standardizes the distribution of the output activations. Although using batch normalization is not crucial in our case, it produces the desired distribution of activations that don't vary too much when the weights change. Said otherwise, it minimizes the internal covariate shift and aids the model to learn faster.

Finally, because we use a different and more complex architecture than that proposed by Yaakoubi et al. \cite{Yaakoubi2019}, our predictor needs much more fine-tuning. Therefore, we use extensive Bayesian Optimization using the Gaussian process (500 iterations) to find a good configuration of hyperparameters. Note that one can use random search or grid search to fine-tune. But, Bayesian optimization was shown to provide much better results for hyperparameter tuning \cite{snoek2012practical,dahl2013improving}.

\subsection{Transformed input}\label{sec:_Transf_Input}

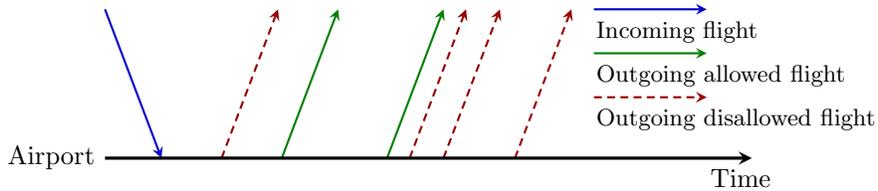
\begin{figure}[H]
\centering
\begin{tikzpicture}[yscale=0.9]

\draw[thick, blue!80!black,-stealth] (0,2.2)--(0.75,0);
\draw[thick, red!60!black,-stealth, densely dashed] (1.55,0)--(2.3,2.2);
\draw[thick, green!50!black,-stealth, xshift=0.8cm] (1.55,0)--(2.3,2.2);
\draw[thick, green!50!black,-stealth, xshift=2.2cm] (1.55,0)--(2.3,2.2);
\draw[thick, red!60!black,-stealth, densely dashed, xshift=2.5cm] (1.55,0)--(2.3,2.2);
\draw[thick, red!60!black,-stealth, densely dashed, xshift=2.95cm] (1.55,0)--(2.3,2.2);
\draw[thick, red!60!black,-stealth, densely dashed, xshift=3.9cm] (1.55,0)--(2.3,2.2);
\draw[thick, blue!80!black,-stealth] (6.5,2.2)node[yshift=-0.3cm,xshift=-0.1cm, black,font=\small, anchor=west]{Incoming flight}--(8,2.2);
\draw[thick, green!50!black,-stealth, yshift=-0.65cm] (6.5,2.2)node[yshift=-0.3cm,xshift=-0.1cm, black,font=\small, anchor=west]{Outgoing allowed flight}--(8,2.2);
\draw[thick, red!60!black,-stealth, densely dashed, yshift=-1.3cm] (6.5,2.2)node[yshift=-0.3cm,xshift=-0.1cm, black,font=\small, anchor=west]{Outgoing disallowed flight}--(8,2.2);
\draw[very thick, -stealth] (0,0)node[left]{Airport}--(8.6,0)node[below, xshift=-0.15cm]{Time};
\end{tikzpicture}
\caption{Illustration of different feasible possibilities for an incoming flight}
\label{Fig:Incoming_Flight_Scenario}
\end{figure}

Given that the aircrew arrived at a specific airport at a given time, we can use \emph{a priori} knowledge to define which flights are possible. For example, as shown in Figure \ref{Fig:Incoming_Flight_Scenario}, it is not possible to make a flight that starts ten minutes after the arrival, nor is it possible five days later. Furthermore, it is rare that the type of aircraft changes between flights since each aircrew is formed to use one or two types of aircraft at most. The reader is referred to~\cite{Kasirzadeh2017} for further details on the likelihood of these scenarios. Note that these simplistic conditions do not depend on the airline company, are applicable to various airline industry problems, and represent \textbf{soft feasibility conditions}. They are not the same conditions that are used by the GENCOL optimizer, which uses the complete list of collective agreements and the exact cost function.

In our work, we use the following conditions, that need to be always satisfied for the next flight performed by the crew: 
\begin{itemize}
\itemsep-0.25mm
    \item The departure time of the next flight should follow the arrival time of the previous flight to the connecting city;
    \item The departure time of the next flight should not exceed 48 hours following the arrival time of the previous flight to the connecting city;
    \item The departure city of the next flight should be identical to the connecting city in the previous flight;
    \item The aircraft type should be the same. Indeed, crew scheduling is separable by crew category and aircraft type or family~\cite{Kasirzadeh2017}.
\end{itemize}

We can, therefore, design a more useful encoding for the output classes by normalizing their representations across different instances. Given a vertex $v$, we would like to predict which arc is most probably after $v$ in a pairing that contains $v$. One of the main difficulties, when one wants to use NNs to predict relevant information on a combinatorial optimization problem, is that a NN accepts only standardized inputs (i.e., vectors in $\mathbb{R}^d$ with $d$ fixed). In contrast, an instance of combinatorial optimization problems is generally not a vector in $\mathbb{R}^d$ with $d$ fixed. Here, the number of arcs outgoing from $v$ depends on $v$. We sparsify the graph to overcome this difficulty. Sparsifying the graph is indeed a generic technique to scale-up a column generation for a vehicle or crew scheduling problem without deteriorating too much the quality of the solution returned. It consists of removing most of the arcs of the graph and keeping only those that are the best candidates to be in the solution. For the CPP, the most relevant arcs are those corresponding to connections such that our masking constraints are met. Previous work on weekly CPP \cite{Yaakoubi2019} has shown that keeping only the 20 most probable outgoing arcs of each vertex leads to excellent solutions. We exploit this technique to obtain a standardized input of the NN: Indeed, after sparsification, all vertices have the same maximal number of outgoing arcs. A vector of $d$ features for each of these arcs will therefore always have the same dimension, $20d$, and can, therefore, be used as an input of the NN (which predicts, for a pairing that contains a vertex $v$, the most probable arc after
$v$.)

We use the embedding layer described earlier to construct a feature representation for each of these 20 flights. Then, we concatenate them to have a matrix $n_{d}\,\times\,$20 input for the next layers, where $n_{d}$ is the embedded representation of information on one flight. Consequently, we not only reduce the number of possible next flights but also construct a similarity-based input, where the neighboring factors have similar features, which in turn allows the usage of CNNs. The intuition here is that we can consider each next flight as a different time step, enabling the use of convolutional architecture across time~\cite{Debayle2018, Borovykh2017}.

\subsection{Cluster construction}\label{sec:Cluster_Construction}

Upon the finalization of the flights-connection prediction model training, we can use the same architecture to solve two other prediction problems on the test set (50 000 flights): (i) predict if each of the scheduled flights is the beginning of a pairing or not; and (ii) predict whether each flight is performed after a layover or not. In reality, the three predictors share the same representation. To solve these independent classification problems, we sum the three prediction problems' cross-entropy losses when learning, therefore performing a multi-output classification.

\paragraph{Start in a base.}While training the NN predictor to recognize whether a flight is the beginning of a pairing or not, it is possible that it misclassifies that a flight departing from a non-base city is indeed the beginning of a pairing. It is imperative to correct such false predictions in order to avoid pairings starting away from the base. Even though it is possible to construct a predictor to which only flights departing from the bases are given, it is more efficient and robust to use all flights in the training step. That way, the predictor learns a better representation of the input.

\paragraph{No Layover below a threshold.} While training the NN predictor to recognize whether there is a layover or not between two flights, and since this decision is independent of finding the next flight, we can use a threshold on the number of hours between the previous and the next flight, below which it does not make sense to make a layover. This threshold should be defined considering previous solutions or by a practitioner.

To build the crew pairing, one can use the following heuristics:
\begin{itemize}
    \item Heuristic 1 (H1): We use a greedy heuristic to build a crew pairing. Specifically, we consider each flight that the model predicts at the beginning of a pairing as a first flight. Given this incoming flight, we predict whether the crew is making a layover or not. In both cases, we consider the incoming flight and predict the next one. The pairing ends when the maximal number of days permitted per pairing is approached.
    We can use the above heuristic to construct a solution for the testing data, obtaining a crew pairing that can be fed as an initial cluster for the solver. Unfortunately, if one flight in the pairing is poorly predicted, as the flights are predefined, the crew can finish its pairing away from the base. 
    Such pairings are discarded.

    \item Heuristic 2 (H2): Like H1, but we also discard pairings where the predictor abstains. Indeed, instead of taking all the predictions into account in constructing the initial clusters for DCA, one can also discard a percentage of these predictions to enhance accuracy. We consider augmenting our classifier with an ``abstain'' option~\cite{nadeem10a}. Since MPDCA adds in phase $k$ in the sub-problem a constraint forbidding to generate a pairing breaking clusters more than $k$ times, breaking a connection in a cluster takes more time than building one. Therefore, removing low confidence links (flights) using this ``abstain'' option can be advantageous.

    \item Adaptable Heuristic 1 (Adapt-H1) and Adaptable Heuristic 2 (Adapt-H2): Because the monthly problem is solved using a rolling-horizon approach with one-week windows and two days of overlap period, constructing an initial partition for the entire month and using the subset in each window to feed DCA can be a major flaw. Such initial partition will have many inconsistencies with the solution of the previous window, particularly during the overlap period, such as legs belonging to two different clusters.
    We propose to adapt the proposed clusters to the solution of the previous window using the heuristics proposed in Section \ref{sec:Cluster_Construction} to construct the clusters of the current window in accordance with the solution found for the previous window and any inconsistency with the previous window is avoided, so the proposed partition is adapted to the current resolution.
    In addition, instead of only considering the flights that the model predicts at the beginning of a pairing as a first flight, incomplete clusters from the solution of the previous window starting during the overlap period are completed.
    For the next section, we denote by Adapt-H1 and Adapt-H2 the heuristics proposing such adaptation to the partitions produced with H1 and H2.
\end{itemize}

\section{Computational experiments}\label{sec:Comp_Exp}

In this section, we report the results of the computational experiments we conducted using \textit{Baseline} and Commercial-GENCOL-DCA for large scale CPPs. First, we present the instances used for training and testing in Section \ref{sec:Instances_Description}, then the hyperparameters and hardware used in the experimentation in Section \ref{Sec:Par_Setting}. Finally, Sections \ref{Sec:ML_Results} and \ref{sec:OR_Results} report our ML prediction results and the crew-pairing problem resolution using Commercial-GENCOL-DCA.

\subsection{Instances description}\label{sec:Instances_Description}

Our training instances originate from a major airline and consist of six monthly crew pairing solutions for approximately 200 cities and 50 000 flights per month. The test set is a different instance, which is a benchmark used by airlines to decide which commercial solver to use. Each instance contains a one-month flight schedule, crew pairings, as well as a list of airports and bases. This information is used to create input for the learning phase. The dataset consists of approximately 1 000 different source-destination pairs, 2 000 different flight codes and pairings start from 7 different bases.

\subsection{Parameters setting}\label{Sec:Par_Setting}

The parameters setting is relevant to the hyperparameter optimization, evaluation process, and hardware description.
To find the best configuration of hyperparameters, we use Bayesian optimization with $k$-fold cross-validation on the training set to measure the configuration quality. These are presented in Table \ref{Table:ML_Hyperparameters}. We use different months for different folds (6 folds) to simulate the more realistic scenario where we make a prediction over a new period of time.

More specifically, we optimize the hyperparameters listed in Table \ref{Table:ML_Hyperparameters}~\cite{Dernoncourt2016} with an implementation of Gaussian process-based Bayesian optimization provided by the GPyOpt Python library version 1.2.1~\cite{gpyopt2016}. Bayesian optimization constructs a probabilistic model of the function mapping from hyperparameter settings to the model performance and provides a systematic way to explore the space more efficiently~\cite{Hutter2011}. To identify the following best candidate to sample, we choose the point that maximizes an acquisition function. One of the most popular acquisition functions is of the type Expected Improvement ($\mathcal{EI}$), which represents the belief that new trials will improve upon the current best configuration. The one with the highest $\mathcal{EI}$ will be tested directly after. Maximizing $\mathcal{EI}$ gives a clear indicator of the region from which we should sample, in order to gain the maximum information about the location of the global maximum of the model performance function. As mentioned above, one can use random search or grid search to fine-tune. But, Bayesian optimization was shown to provide much better results for hyperparameter tuning \cite{snoek2012practical,dahl2013improving}.

In our approach, the optimization was initialized with 50 random search iterations, followed by up to 450 iterations of standard Gaussian process optimization. Here, the test accuracy is used as the surrogate function and $\mathcal{EI}$ as the acquisition function.

\begin{table*}[htb]
\small
\begin{center}
\caption{Hyperparameters used in optimization}
 \label{Table:ML_Hyperparameters}
\begin{tabular}{l|l|l}

\toprule
\textbf{Parameters} &
\textbf{Search space} &
\textbf{Type} \\
\midrule
Optimizer &
Adadelta; Adam; Adagrad; Rmsprop &
Categorical \\
Learning rate &
0.001, 0.002, \dots, 0.01 &
Float \\
Dimensions of the embeddings &
5, 10, 15, \dots, 50 &
Integer \\
Number of dense layers &
1, 2, 3, 4, 5 &
Integer \\
Neurons per layer &
100, 200, \dots, 1000 &
Integer \\
Dropout rate &
0.1, 0.2, \dots, 0.9 &
Float \\
Convolutional layers &
0, 1, 2, 3 &
Integer \\
Filters $n$ &
50, 100, 250, 500, 1000 &
Integer \\
Filter size $h$ &
3, 4, 5 &
Integer \\
\bottomrule
\end{tabular}
\end{center}
\end{table*}

All experiments were executed on a 40-core machine with 384\,GB of memory. Each method is executed in an asynchronously parallel setup of 2-4 GPUs. That is, it can evaluate multiple models in parallel, with each model on a single GPU. When the evaluation of one model is completed, the methods can incorporate the result and immediately re-deploy the next job without waiting for the others to be finalized. We use four K80 (12\,GB) GPUs with a time allocation of 10 hours. All algorithms were implemented in Python using Keras~\cite{chollet2015keras} and Tensorflow~\cite{tensorflow2015} libraries.

\subsection{Results on Next-Flight-Prediction}\label{Sec:ML_Results}

We perform the Gaussian process to search for the best configuration of hyperparameters.
To warm start the method with initial samples, we first use random search.
After only a few iterations of random search, we are able to get an accuracy of 99.35\%. Then, random search boosts the total return very quickly up to 99.62\% after 18 iterations and thus remains until the end of the random search cycle (iteration number 50). Using Bayesian optimization, we can show that we continuously improve our process of searching for the best configuration of hyperparameters that maximizes the overall return. In our case, we stopped at iteration number 500 with the best architecture providing an accuracy of 99.68\%.

Using the ``abstain'' option, the accuracy increases from 99.62\% to 99.94\%, estimating confidence with dropout: The prediction tasks can be carried $N$ times while applying dropout in the entire layers of the NNs~\cite{Gal2016}, yielding $N$ probability vectors for each test sample. A rough estimate of certainty of prediction is obtained by computing the mean of these $N$ probability vectors and subtracting their component-wise estimated standard deviation (computed from the same $N$ vectors). This gives a lower bound on the certainty of our prediction. If the maximum value of these confidences is too low, we decide to abstain. For the next subsection, we will use a 0.5\% rejection rate (abstention) for Heuristic~2~(H2).

\subsection{Results on crew pairing problems}\label{sec:OR_Results}

Computational results per window are reported in Table \ref{Table:Solver_Results} for all algorithms, namely, Baseline, H1, H2, Adapt-H1, and Adapt-H2. For each window and each algorithm, we provide the LP value at the root node of the search tree N0 (LP-N0), the computational time at N0 (N0 time), the number of fractional variables (\# FV-N0) in the current MP solution at N0, the number of branching nodes resolved (\# Nodes), the best LP value found (Best-LP), the pairing cost of the best feasible solution (INT) and finally the total computational time (T time); times are in seconds. Furthermore, for all ML algorithms, and for LP-N0, Best-LP, and INT, we indicate the relative difference between the result obtained with this algorithm and that with Baseline.

\begin{table}
\centering
\resizebox{\textwidth}{!}{
\setlength{\tabcolsep}{2.5pt}
\begin{tabular}{ccC{0.1mm}ccC{0.1mm}ccC{0.1mm}C{1.3cm}cC{0.1mm}ccC{0.1mm}ccC{0.1mm}cc}
\hline
Win.&Alg. & ~ & LP-N0&Diff. & ~ &\#FV-N0& Diff. & ~ &\#Nodes & Diff. &  ~ & Best-LP&Diff.    & ~ & INT&Diff. & ~ & T time & Diff. \\
 & & ~ & & (\%) & ~ &  & (\%) & ~ & & (\%) &  ~ & & (\%)& ~ &  &(\%)& ~ &(s) & (\%) \\
\cline{1-2}  \cline{4-5}  \cline{7-8}  \cline{10-11} \cline{13-14} \cline{16-17}  \cline{19-20} 
      & Baseline &  & 10122035 &        &  &    2719 &      &  & 159  &  &  & 10025487.73 &        &  & 10276344.14 &        &  & 7891 &        \\
      & H1       &  & 9904828  & -2.15  &  & 2870 & 5.55   &  & 203 & 27.67 &  & 9891390.17  & -1.34  &  & 10136106.73 & -1.36  &  & 10405    &  31.86 \\
1      & H2       &  & 9889525  & -2.30  &  & 3014& 10.85    &  & 249 & 56.60 &  & 9877050.50  & -1.48  &  & 10300447.14 & 0.23   &  & 11455   &  45.17 \\
      & Adapt H1 &  & 9904828  & -2.15  &  & 2870 & 5.55    &  & 203 & 27.67 &  & 9891390.17  & -1.34  &  & 10136106.73 & -1.36  &  & 10691    & 35.48  \\
      & Adapt H2 &  & 9889525  & -2.30  &  & 3014 & 10.85    &  & 249 & 56.60 &  & 9877050.50  & -1.48  &  & 10300447.14 & 0.23   &  & 13642    & 72.88  \\
       &          &  &          &        &  &  &&       &  &      &  &             &        &  &             &        &  &        &        \\
      & Baseline &  & 11396498 &        &  & 2519&    &  & 162 &  &  & 11225022.17 &        &  & 11501016.42 &        &  & 27778     &        \\
      & H1       &  & 10589590 & -7.08  &  & 4771 & 89.40   &  & 446 & 175.30 &  & 10426771.03 & -7.11  &  & 13080746.04 & 13.74  &  & 63186     & 127.47 \\
2      & H2       &  & 10528757 & -7.61  &  & 5048 & 100.40   &  & 381 & 135.19 &  & 10386235.55 & -7.47  &  & 11133714.80 & -3.19  &  & 63462    & 128.46 \\
      & Adapt H1 &  & 10319719 & -9.45  &  & 5848 & 132.16   &  & 419 & 158.64 &  & 10215252.88 & -9.00  &  & 10865255.80 & -5.53  &  & 94004     & 238.41 \\
      & Adapt H2 &  & 10271980 & -9.87  &  & 6182 & 145.42   &  & 492 & 203.70 &  & 10188780.86 & -9.23  &  & 10990621.86 & -4.44  &  & 126761    & 356.33 \\
       &          &  &          &        &  & &        &  &  &    &  &             &        &  &             &        &  &        &        \\
      & Baseline &  & 10740127 &        &  & 2330 &    &  & 177 &  &  & 10641496.00 &        &  & 11107990.12 &        &  & 30421     &        \\
      & H1       &  & 10590614 & -1.39  &  & 5340 & 129.18  &  & 366 & 106.78 &  & 10439285.84 & -1.90  &  & 10926909.75 & -1.63  &  & 67217     & 120.96 \\
3      & H2       &  & 9850067  & -8.29  &  & 4561& 95.75   &  & 307 & 73.45 &  & 9659421.20  & -9.23  &  & 10272961.85 & -7.52  &  & 52748    & 73.39  \\
      & Adapt H1 &  & 9596326  & -10.65 &  & 4838 & 107.64  &  & 294 & 66.10 &  & 9432461.69  & -11.36 &  & 10051850.40 & -9.51  &  & 53736     & 76.64  \\
      & Adapt H2 &  & 9685782  & -9.82  &  & 4574 & 96.31  &  & 308 & 74.01 &   & 9483718.96  & -10.88 &  & 10160044.85 & -8.53  &  & 68597     & 125.49 \\
       &          &  &          &        &  &  &        &  &    &  &  &             &        &  &             &        &  &        &        \\
      & Baseline &  & 9764727  &        &  & 2606  &   &  & 229 &  &  & 9591592.13  &        &  & 9968081.73  &        &  & 33898     &        \\
      & H1       &  & 9164173  & -6.15  &  & 3808 & 46.12  &  & 314 & 37.12 &  & 9015056.00  & -6.01  &  & 13618635.11 & 36.62  &  & 37142     & 9.57   \\
4      & H2       &  & 8007573  & -17.99 &  & 5264 & 102.00  &  & 358 & 56.33 &  & 7872923.59  & -17.92 &  & 9619055.54  & -3.50  &  & 79271    & 133.85 \\
      & Adapt H1 &  & 8063084  & -17.43 &  & 4763 & 82.77  &  & 394 & 72.05 &  & 7868177.74  & -17.97 &  & 8791799.94  & -11.80 &  & 56291     & 66.06  \\
      & Adapt H2 &  & 7574006  & -22.44 &  & 6473 & 148.39  &  & 463 & 102.18 &  & 7516289.41  & -21.64 &  & 8333643.43  & -16.40 &  & 152781    & 350.71 \\
       &          &  &          &        &  &         &  &      &  &             &        &  &             &        &  &        &        \\
      & Baseline &  & 8095063  &        &  & 3150 &   &  & 188 &  &  & 7899149.01  &        &  & 8102703.93  &        &  & 35948     &        \\
      & H1       &  & 9375047  & 15.81  &  & 3571 & 13.37  &  & 303 & 61.17 &  & 9193080.40  & 16.38  &  & 10730905.63 & 32.44  &  & 34123     & -5.08  \\
5      & H2       &  & 6305624  & -22.11 &  & 4558 & 44.70   &  & 362 & 92.55 &  & 6146413.51  & -22.19 &  & 6746656.99  & -16.74 &  & 50138    & 39.47  \\
      & Adapt H1 &  & 6076461  & -24.94 &  & 5333 & 69.30  &  & 417 & 121.81 &  & 5953932.68  & -24.63 &  & 6453605.80  & -20.35 &  & 74622     & 107.58 \\
      & Adapt H2 &  & 5706073  & -29.51 &  & 6008 & 90.73  &  & 480 & 155.32 &  & 5646359.35  & -28.52 &  & 6464805.10  & -20.21 &  & 133395    & 271.78 \\
       &          &  &          &        &  &    &     &  &      &  &    &         &        &  &             &        &  &        &        \\
      & Baseline &  & 6778925  &        &  & 2513 &    &  & 187 &  &  & 6610562.32  &        &  & 6939096.41  &        &  & 29368     &        \\
      & H1       &  & 6611841  & -2.46  &  & 3960 & 57.58   &  & 324 & 73.26 &  & 6432282.33  & -2.70  &  & 8915306.53  & 28.48  &  & 39626     & 34.93  \\
6      & H2       &  & 4905952  & -27.63 &  & 4882 & 94.27  &  & 422 & 125.67 &  & 4814868.16  & -27.16 &  & 5297148.54  & -23.66 &  & 61760    & 110.30 \\
      & Adapt H1 &  & 4763520  & -29.73 &  & 4940 & 96.58  &  & 318 & 70.05 &  & 4697990.54  & -28.93 &  & 4947363.87  & -28.70 &  & 55263     & 88.17  \\
      & Adapt H2 &  & 4763509  & -29.73 &  & 5713 & 127.34  &  & 437 & 133.69 &  & 4726686.40  & -28.50 &  & 5114636.73  & -26.29 &  & 94092     & 220.39 \\
       &          &  &          &        &  &     &  &  &  &      &  &             &        &  &             &        &  &        &        \\
\hline
       &          &  &          &        &  &     &  &  &  &      &  &             &        &  &             &        &  &        &        \\
      & Baseline  &  & 9482896 &         &  & 2640 &    & &  183 &   &  & 9332218.23  &        &  & 9649205.46  &        &  & 27551  &        \\
      & H1        &  & 9372682 & -0.57   &  & 4053 & 53.52   & &  326 & 78.14  &  & 9232977.63  & -0.45  &  & 11234768.30 & 18.05  &  & 41950  & 52.26  \\
Mean  & H2        &  & 8247916 & -14.32  &  & 4555 & 72.54   & &  346 & 89.07  &  & 8126152.09  & -14.24 &  & 8894997.48  & -9.06  &  & 53139  & 92.88  \\
      & Adapt H1  &  & 8120656 & -15.73  &  & 4765 & 80.49   & &  340 & 85.79 &  & 8009867.62  & -15.54 &  & 8540997.09  & -12.88 &  & 57435  & 108.47 \\
      & Adapt H2  &  & 7981813 & -17.28  &  & 5327 & 101.78   & &  404 & 120.76 &  & 7906480.91  & -16.71 &  & 8560699.85  & -12.61 &  & 98211  & 256.47 \\
\hline
\end{tabular}}
\caption{Computational results per window}
\label{Table:Solver_Results}
\end{table}

Overall, observe first that the versions with ML (and practically Adapt-H1) produce more stable results in each weekly window than Baseline, which does not produce improvements in each window. In some cases in Baseline, the heuristic branch-\&-bound did not find an integer solution good enough, and in some cases, the LP optimization was poor because the initial information was poor, and the explored neighborhood was too small to reach a good solution. These weaknesses were solved in Commercial-GENCOL-DCA with ML through the improvements to branch-\&-bound in box 8 and the dynamic clusters described in Sections \ref{sec:imp_rolling_horizon} and \ref{sec:Cluster_Construction}.

We start by comparing Baseline, H1, and H2.
Observe first that H2 gives lower LP-N0 and Best-LP values with an average reduction factor of 14.32\% and 14.24\%, respectively, compared to Baseline, while H1 gives lees significant reductions in LP-N0 and Best-LP values with a reduction factor of only 0.57\% and 0.45\%, respectively. The same can be observed for the cost of the best feasible solution (INT), where H2 has an average reduction factor of 9.06\% while H1 gives worse results with an increase factor of 18.05\%. Note that not all lower bounds at the root node N0 are equal, depending on the methodology used. Fluctuations in LP-N0 values assess the industrial and, therefore, particularly challenging nature of the problem at hand to even produce initial starting solutions of good quality. This is due to the fact that the root node is not solved to optimality and, therefore, that even optimizing the linear relaxation is difficult for a monthly CPP of 50 000 flights. H1 does not perform well because the proposed clusters are not adapted to the solution of the previous window found by the optimizer, while this problem seems to be partially avoided when using the abstention method. This is explained by the ability of H2 to discard poorly predicted clusters by using the ``abstain'' option. Since the optimizer takes more time to break a connection in a cluster (links) than to build one, removing low confidence links (flights) using this ``abstain'' option can be advantageous.

On the other hand, note that for H1 and H2, the average total computational times per window are between 9.57\% and 133.85\% larger than those of Baseline. This time increase is due to the large number of fractional variables at the root node of the search tree with an increase factor between 5.55\% and 129.18\%. This is explained by the fact that, when base constraints are restrictive, the root node solutions contain a more significant number of fractional-valued pairing variables in order to split the worked time between the bases evenly. This causes an increase in the number of branching nodes required to obtain a good integer solution. Indeed, H1 and H2 present an increase factor in the number of branching nodes between 27.67\% and 175.30\%.

Next, we compare Baseline, Adapt-H1, and Adapt-H2. For the root node (N0), observe that both adaptation-based heuristics give lower LP-N0 values for all windows providing an average reduction factor of 15.73\% and 17.28\%, respectively. Likewise, both heuristics provide better Best-LP values with an average reduction factor of 15.54\% and 16.71\% and better feasible solutions with a reduction factor of 12.88\% and 12.61\%. This is explained by the ability of Adapt-H1, and Adapt-H2 to propose custom-made clusters in an online manner adapted to the solution of the previous window, completing clusters that start in the overlap period and proposing new unseen clusters for the non-overlap period.

The computational times are high because we explore a larger neighborhood using $p=.6$ (Box 6). We run tests with $p=.3$ and $p=.6$, and we select $p=.6$ because the objective was to obtain the best solutions. The solution time will be adjusted later in the industrial environment on workstations using many processors.

On the other hand, note that the average computational time for Adapt-H1 is similar to H2 while that of Adapt-H2 is, on average, 76.46\% larger. This is due to the larger number of nodes caused by the larger number of fractional variables at the root node of the search tree. This can be explained by the fact that Adapt-H2 discards between 30 and 50 clusters per window, providing fewer clusters than Adapt-H1. Therefore, the adaptation scheme is capable of proposing suitable clusters and the shifting scheme to use the next available flight if the predicted next flight is covered by another crew makes the abstention option unnecessary. The computational times are high because we explore larger neighborhood using $p=.6$ (Box 6 in Figure \ref{fig:new_algo_spp}). We run tests with $p=.3$ and $p=.6$, and we select $p=.6$ because the objective was to obtain the best solutions. The solution time will be adjusted later in the industrial environment on workstations using many processors.
Note also that the dynamic strategies permit large improvements of the LP bound with a relatively small number of fractional variables.

Computational results on monthly solutions are reported in Table \ref{Table:Computational_Results}. We report the solution cost, the cost of global constraints, the number of deadheads, and the total computational time (T time). The solution cost is composed of the cost of pairings and the cost of global constraints. The cost of a pairing approximates the salary of its crew as well as other expenditures, such s hotel costs, per diem rates, and deadheads. The cost of global constraints is a penalty for soft base constraints. The base constraints aim at distributing the workload fairly amongst the bases proportionally to the personnel available at each base. For all heuristics algorithms, we also indicate the relative difference between the result obtained with this algorithm and that with Baseline.

\begin{table}[htb]
\centering
\resizebox{\textwidth}{!}{
\setlength{\tabcolsep}{2pt}
\begin{tabular}{cC{0.1mm}ccC{0.1mm}ccC{0.1mm}ccC{0.1mm}c}
\hline
 && Solution cost & Diff. vs. Baseline && Cost of & Diff. vs. Baseline && Number of  & Diff. vs. Baseline && T time \\
 &&           & (\%) &&     global constraints       & (\%) &&     deadheads   & (\%)     && (hours:minutes)       \\
 \cline{1-1} \cline{3-4} \cline{6-7} \cline{9-10} \cline{12-12}
GENCOL init &  & 30 681 120.5              &       &  & 9 465 982.28         &        &  &  1725   &      &&      \\
Baseline     &  & 20 639 814.6   & -32.73 \scriptsize{(vs. GENCOL init)} &  & 2 127 086.77         & -77.53 \scriptsize{(vs. GENCOL init)}  && 992 & -42.49 \scriptsize{(vs. GENCOL init)} && 45:55 \\
H1           &  & 21 118 006.16  & 2.32  &  & 2 202 610.31         & 3.55   &  & 1136     &     14.52  && 69:55   \\
H2           &  & 19 235 343.4   & -6.80 &  & 642 599.29          & -69.79 &  & 1059      &     6.75   && 88:34 \\
Adapt-H1     &  & 18 881 977.89  & -8.52 &  & 465 687.94          & -78.11 &  & 1014      &     2.22   && 95:43 \\
Adapt-H2     &  & 19 104 804.62  & -7.44 &  & 490 787.75          & -76.93 &  & 1097      &     10.58  && 163:41  \\
\hline
\end{tabular}}
\caption{Computational results on monthly solution}
\label{Table:Computational_Results}
\end{table}

We start by comparing GENCOL init and Baseline. Observe first that a significantly better solution was found using Baseline, compared to the initial solution GENCOL init. Indeed, the solver significantly reduced both the solution cost and the cost of global constraints with a reduction factor of 32.73\% and 77.53\%, respectively, while reducing the number of deadheads by 42.49\%. This attests to the optimizer's capacity to tackle larger windows finding better solutions and improving industrial-scale solutions.

Next, we compare Baseline, H1, and H2. While the solution cost and the cost of global constraints found with H1 are slightly worse than Baseline with an increase factor of 2.32\% and 3.55\%, H2 outperforms Baseline reducing the solution cost by 6.8\%. Furthermore, H2 reduced the cost of global constraints by 69.79\%, which supports and justifies the large number of fractional variables at the root node of the search tree and the number of nodes, yielding larger computational times. We believe that this trade-off is acceptable since the improvement in the cost of global constraints is significant and that the larger computational times are due to the tight constraints on the number of worked hours per base. Relaxing these constraints may yield better results than Baseline while reducing the computational time. Finally, note that the poor results of H1 and good results of H2 are explained by the ability of H2 to tackle and revise poor predictions and discard poorly constructed clusters.

Then, we compare Baseline, Adapt-H1, and Adapt-H2. The solutions found by both heuristics present better statistics than Baseline, H1, and H2. Adapt-H1 yields better solutions than any other heuristic, with a reduction factor in the solution cost and cost of global constraints of 8.52\% and 78.11\%, respectively. Adapt-H2 presents similar results while providing a reduction factor of 7.44\% and 76.93\%. It is also worth noting that, for all heuristics, the number of deadheads used is slightly larger with an increase factor between 2.22\% and 14.52\%, compared to Baseline. The cost of deadheads is accounted for in the solution cost. Because the solution cost is a multi-objective function, we believe that using slightly more deadheads permitted to get better solutions, enhancing both the solution cost and the cost of global constraints.

\section{Conclusion} \label{sec:Conclusion}

In this paper, we present Commercial-GENCOL-DCA, an improved implementation of \textit{Baseline} \cite{desaulniers2019} that relies on column generation and constraint aggregation: DCA, MPDCA, and IPS. On the one hand, DCA only incorporates set partitioning constraints. On the other hand, IPS is capable of handling additional linear constraints but does not provide fast techniques for identifying compatible columns. \textit{Baseline} constructs the complementary problem of IPS by working on clusters and flights with the network techniques of DCA to identify compatible and incompatible variables. It also combines some new techniques of partial pricing and Branch-\&-Bound to solve a large-scale problem where DCA, MPDCA, and IPS have not been applied before: monthly CPP with complex industrial constraints.

Commercial-GENCOL-DCA improves upon \textit{Baseline} by using a dynamic control strategy and a variable aggregation and disaggregation strategy reducing the size of the problem without negative aspects on the number of column generation iterations and the complexity of branching decisions. In addition, the column generation solver is modified to take strong advantage of an initial solution and clusters of flights with a large probability to be consecutive in a solution to speed up the column generation algorithm.

We developed ML-based heuristics capable of constructing adapted initial clusters for the optimizer, taking into account multiple past solutions. This work is the first attempt to embed into a column generation framework recently developed ML methods. We also proposed an adaptation mechanism to propose clusters in an online manner, taking into account the solution of the previous window.

We compared the performance of the \textit{Baseline} solver \cite{desaulniers2019} (using a standard initial solution as clusters) with Commercial-GENCOL-DCA (using clusters proposed by ML-based heuristics). The main computational results show that Commercial-GENCOL-DCA yields better results than \textit{Baseline} using a prototype rolling-horizon approach with narrow windows. In addition, ML-based heuristics, taking advantage of abstention or adaptation, yield better results with significantly smaller costs reducing by 8.52\% the solution cost, and by 78.11\% the cost of global constraints.

We believe that the proposed solution is able to handle larger problems as well as learn from multiple airline companies to construct initial clusters for a new company since it does not use flight codes in any part of the pre-processing, learning or prediction process. We also believe that the combination of ML and optimization with constraint aggregation can easily be adapted to other types of optimization problems, such as railway or bus shift scheduling. Indeed, it has been observed numerous times that warm-starting column generation with an optimal integer basis can even be counterproductive to the overall branch-\&-price. This paper presents results illustrating that algorithms using constraint aggregation (DCA, MPDCA, and IPS) can achieve high acceleration with good initial information. It is useful information for the large community using column generation.

A potential improvement to our work is to propose a ML predictor that is capable of providing a monthly solution that can be used both as initial clusters and as an initial solution. Unlike our NN predictor, such a ML predictor cannot be greedy and must take into consideration the graph structure to maximize the feasibility of the proposed solution.

Finally, our long-term goal is to develop efficient new learning techniques that could handle and learn from the flight-based network structure, where nodes correspond to time-space coordinates and arcs represent tasks performed by crew members (legs, deadheads, connections, rests, etc.) capable of incorporating global and local constraints in the ML-predictor learning process.

\section{Acknowledgements}
The authors thank the anonymous reviewers for their valuable comments that improved the quality of this work. This work was supported by IVADO and a Collaborative Research and Development Grant from the Natural Sciences and Engineering Research Council of Canada (NSERC) and AD OPT, a division of IBS Software. The authors would like to thank these organizations for their support and confidence.

\bibliography{bibliography}

\end{document}